\documentclass{article}

\usepackage[preprint,nonatbib]{neurips_2021}

\usepackage[utf8]{inputenc} 
\usepackage[T1]{fontenc}    
\usepackage{hyperref}       
\usepackage{url}            
\usepackage{booktabs}       
\usepackage{nicefrac}       
\usepackage{microtype}      
\usepackage{xcolor}         
\usepackage{graphicx}
\usepackage{enumitem} 
\usepackage{subcaption}  
\usepackage{amssymb} 
\usepackage{algorithm}
\usepackage{algorithmic}

\usepackage{amsmath,amsfonts,bm,amsthm}

\newtheorem{theorem}{Theorem}
\newtheorem{definition}{Definition}
\newtheorem{corollary}{Corollary}

\def\eqref#1{equation~\ref{#1}}

\def\floor#1{\lfloor #1 \rfloor}
\def\1{\bm{1}}

\DeclareMathAlphabet{\mathsfit}{\encodingdefault}{\sfdefault}{m}{sl}
\SetMathAlphabet{\mathsfit}{bold}{\encodingdefault}{\sfdefault}{bx}{n}

\usepackage{hyperref}
\usepackage{multirow}
\newcommand*\samethanks[1][\value{footnote}]{\footnotemark[#1]}

\title{Speedy Performance Estimation for Neural Architecture Search}

\author{%
  Binxin Ru \thanks{These authors contributed equally.} \\
  Department of Engineering Science\\
  University of Oxford \\
  \texttt{robin@robots.ox.ac.uk} \\
   \And
  Clare Lyle \samethanks \\
  Department of Computer Science\\
  University of Oxford \\
  \texttt{clare.lyle@univ.ox.ac.uk} \\
   \AND
  Lisa Schut  \\
  Department of Computer Science\\
  University of Oxford \\
  \texttt{lisa.schut@cs.ox.ac.uk} \\
  \And
  Miroslav Fil \\
  Department of Computer Science\\
  University of Oxford \\
  \texttt{miroslav.fil@seh.ox.ac.uk} \\
  \AND
  Mark van der Wilk \\
  Department of Computing \\
  Imperial College London \\
  \texttt{m.vdwilk@imperial.ac.uk} \\
  \And
  Yarin Gal  \\
  Department of Computer Science\\
  University of Oxford \\
  \texttt{yarin@cs.ox.ac.uk} \\
}

\begin{document}

\maketitle

\begin{abstract}
   Reliable yet efficient evaluation of generalisation performance of a proposed architecture is crucial to the success of neural architecture search (NAS). Traditional approaches face a variety of limitations: training each architecture to completion is prohibitively expensive, early stopped validation accuracy may correlate poorly with fully trained performance, and model-based estimators require large training sets. We instead propose to estimate the final test performance based on a simple measure of training speed. Our estimator is theoretically motivated by the connection between generalisation and training speed, and is also inspired by the reformulation of a PAC-Bayes bound under the Bayesian setting. Our model-free estimator is simple, efficient, and cheap to implement, and does not require hyperparameter-tuning or surrogate training before deployment. We demonstrate on various NAS search spaces that our estimator consistently outperforms other alternatives in achieving better correlation with the true test performance rankings. We further show that our estimator can be easily incorporated into both query-based and one-shot NAS methods to improve the speed or quality of the search.
\end{abstract}

\section{Introduction}

Reliably estimating the generalisation performance of a proposed architecture is crucial to the success of Neural Architecture Search (NAS) but has always been a major bottleneck in NAS algorithms \cite{elsken2018neural}. The traditional approach of training each architecture for a large number of epochs and evaluating it on validation data (\emph{full training}) provides a reliable performance measure, but requires prohibitively large computational resources on the order of thousands of GPU days \cite{ZophLe17_NAS,Real2017_EvoNAS,zoph2018learning,real2019regularized,elsken2018neural}. This motivates the development of methods for speeding up performance estimation to make NAS practical for limited computing budgets. 

A popular simple approach is \emph{early-stopping}, which offers a low-fidelity approximation of generalisation performance by training for fewer epochs \cite{li2016hyperband,falkner2018bohb,Li2019_random}. 
However, if we stop training early and evaluate the model on validation data, the relative performance ranking may not correlate well with the performance ranking of the full training \cite{zela2018towards}, i.e. the test performance of the model after the entire trainin budget has been used.
Another line of work focuses on \emph{learning curve extrapolation} \cite{domhan2015speeding,klein2016learning,baker2017accelerating}, which trains a surrogate model to predict the final generalisation performance based on the initial learning curve and/or meta-features of the architecture. However, the training of the surrogate often requires hundreds of fully evaluated architectures to achieve satisfactory extrapolation performance and the hyper-parameters of the surrogate also need to be optimised. Alternatively, the idea of \emph{weight sharing} is adopted in one-shot NAS methods to speed up evaluation \cite{Pham2018_ENAS,Liu2019_DARTS,Xie19_SNAS}. Despite leading to significant cost-saving, weight sharing heavily underestimates the true performance of good architectures and is unreliable in predicting the relative ranking among architectures \cite{Yang2020NASEFH,Yu2020Evaluating}. A more recent group of estimators claim to be zero-cost \cite{mellor2020neural, abdelfattah2021zerocost}. Yet, their performance is often not competitive with the state of the art, inconsistent across tasks and cannot be further improved with additional training budgets. 

In view of the above limitations, we propose a simple model-free method, Training Speed Estimation (TSE), which provides a reliable yet computationally cheap estimate of the generalisation performance ranking of architectures. Our method is inspired by recent empirical and theoretical results linking training speed and generalisation \cite{hardt2016train, lyle2020} and measures the training speed of an architecture by summing the training losses of the commonly-used SGD optimiser during training. We empirically show that our estimator can outperform strong existing approaches to predict the relative performance ranking among architectures, and can remain effective for a variety of search spaces and datasets. Moreover, we verify its usefulness under different NAS settings and find it can speed up query-based NAS approaches significantly as well improve the performance of one-shot and differentiable NAS.

\section{Method}\label{sec:method}\label{sec:tse}

\paragraph{Motivation}
The theoretical relationship between training speed and generalisation is described in a number of existing works. Stability-based generalisation bounds for SGD \cite{hardt2016train, liu2017algorithmic} bound the generalisation gap of a model based on the number of optimisation steps used to train it. These bounds predict that models which train faster obtain a lower worst-case generalisation error.
In networks of sufficient width, a neural-tangent-kernel-inspired complexity measure can bound both the worst-case generalisation gap and the rate of convergence of (full-batch) gradient descent \cite{arora2019fine, cao2019}. 
However, these bounds cannot distinguish between models that are trained for the same number of steps but attain near-zero loss at different rates as they ignore the training trajectory.  

We instead draw inspiration from another approach which incorporates properties of the trajectory taken during SGD, seen in the information-theoretic generalisation bounds of \cite{negreainformation} and \cite{neu2021informationtheoretic}. These bounds depend on the variance of gradients computed during training, a quantity which provides a first-order approximation of training speed by quantifying how well gradient updates computed for one minibatch generalize to other points in the training set.
In view of this link, the empirical and theoretical findings in recent work \cite{fort2019stiffness, smith2021origin}, which show a notion of the variance of gradients over the training data is correlated with generalisation, become another piece of supporting evidence for training speed as a measure of generalisation.

Finally, \cite{lyle2020} prove that, in the setting of linear models and infinitely wide deep models performing Bayesian updates, the marginal likelihood, which is a theoretically-justified tool for model selection in Bayesian learning, can be bounded by a notion of training speed. This notion of training speed is defined as a sum of negative log predictive likelihoods, terms that resemble the losses on new data points seen by the model during an online learning procedure. Maximising Bayesian marginal likelihood is also equivalent to minimising the PAC-Bayesian bound on the generalisation error as shown by \cite{germain2016pac}. In particular, this suggests that using the TSE approach for model selection in Bayesian models is equivalent to minimizing an estimate of a PAC-Bayes bound. Because the NAS settings we consider in our experiments use neural networks rather than Bayesian models and due to space constraints, we defer the statement and proof of this result to Appendix A.


\paragraph{Training Speed Estimation}
The results described above suggest that leveraging a notion of training speed may benefit model selection procedures in NAS. Many such notions exist in the generalisation literature: \cite{jiang2020fantastic} count the number of optimisation steps needed to attain a loss below a specified threshold, while \cite{hardt2016train} consider the total number of optimisation steps taken. Both measures are strong predictors of generalisation after training, yet neither is suitable for NAS, where we seek to stop training as early as possible if the model is not promising.

We draw inspiration from the Bayesian perspective and the PAC-Bayesian bound discussed above and present an alternate estimator of training speed that amounts to the area under the model's training curve. Models that train quickly attain a low loss after few training steps, and so will have a lower area-under-curve than those which train slowly. This addresses the shortcomings of the previous two methods as it is able to distinguish models both early and late in training. 

\begin{definition}[Training Speed Estimator]
Let $\ell$ denote a loss function, $f_\theta(\mathbf{x})$ the output of a neural network $f$ with input $\mathbf{x}$ and parameters $\theta$, and let $\theta_{t, i}$ denote the parameters of the network after $t$ epochs and $i$ minibatches of SGD. After training the network for $T$ epochs\footnote{$T$ can be far from the total training epochs $T_{end}$ used in complete training}, we sum the training losses collected so far to get the following \emph{Training Speed Estimate} (\textbf{TSE}):
\begin{equation}
	\mathrm{TSE }= \sum^T_{t=1} \left[ \frac{1}{B} \sum^B_{i=1} \ell \left( f_{\theta_{t, i}}(\mathbf{X}_i), \mathbf{y}_i \right) \right]
\end{equation}
where $l$ is the training loss of a mini-batch $(\mathbf{X}_i, \mathbf{y}_i)$ at epoch $t$ and $B$ is the number of training steps within an epoch.
\end{definition}

This estimator weights the losses accumulated during every epoch equally. However, recent work suggests that training dynamics of neural networks in the very early epochs are often unstable and not always informative of properties of the converged networks \cite{lewkowycz2020large}. Therefore, we hypothesise that an estimator of network training speed that assigns higher weights to later epochs may exhibit a better correlation with the true generalisation performance of the final trained network. On the other hand, it is common for neural networks to overfit on their training data and reach near-zero loss after sufficient optimisation steps, so attempting to measure training speed \textit{solely} based on the epochs near the end of training will be difficult and likely suffer degraded performance on model selection.


To verify whether it is beneficial to ignore or downplay the information from early epochs of training, we propose two variants of our estimator. The first, \textbf{TSE-E}, treats the first few epochs as a burn-in phase for $\theta_{t,i}$ to converge to a stable distribution $P(\theta)$ and starts the sum from epoch $t=T-E+1$ instead of $t=1$. In the case where $E=1$, we start the sum at $t=T$ and our estimator corresponds to the sum over training losses within the most recent epoch $t=T$.
\begin{equation*}
    \text{TSE-E} = \sum^T_{t=T-E+1} \left[ \frac{1}{B} \sum^B_{i=1} \ell \left( f_{\theta_{t, i}}(\mathbf{X}_i), \mathbf{y}_i \right) \right], \hspace{10pt} \text{TSE-EMA} = \sum^T_{t=1} \gamma^{T-t}  \left[ \frac{1}{B} \sum^B_{i=1} \ell \left( f_{\theta_{t, i}}(\mathbf{X}_i), \mathbf{y}_i \right) \right]
\end{equation*}

The second, \textbf{TSE-EMA}, does not completely discard the information from the early training trajectory but takes an exponential moving average of the sum of training losses with $\gamma=0.9$, thus assigning higher weight to the sum of losses obtained in later training epochs.

We empirically show in Section \ref{subsec:comparison} that our proposed TSE and its variants (TSE-E and TSE-EMA), despite their simple form, can reliably estimate the generalisation performance of neural architectures with a very small training budget, can remain effective for a large range of training epochs, and are robust to the choice of hyperparameters such as the summation window $E$ and the decay rate $\gamma$. However, our estimator is \textit{not} meant to replace the validation accuracy at the end of training or when the user can afford large training budget to sufficient train the model. In those settings, validation accuracy remains as the gold standard for evaluating the true test performance of architectures. Ours is just a speedy performance estimator for NAS, aimed at giving an indication early in training about an architecture's generalisation potential under a fixed training set-up.

Our choice of using the training loss, instead of the validation loss, to measure training speed is an important component of the proposed method. While it is possible to formulate an alternative estimator, which sums the validation losses of a model early in training, this estimator would no longer be measuring \textit{training speed}. In particular, such an estimator would not capture the generalisation of gradient updates from one minibatch to later minibatches in the data to the same extent as TSE does. 
Indeed, we hypothesise that once the optimisation process has reached a local minimum, the sum over validation losses more closely resembles a variance-reduction technique that estimates the expected loss over parameters sampled via noisy SGD steps around this minimum.
We show in Figure~\ref{fig:baseline_compare} and Appendix C that our proposed sum over training losses (TSE) outperforms the sum over validation losses (SoVL) in ranking models in agreement with their true test performance. 

\section{Related Work}
Various approaches have been developed to speed up architecture performance estimation, thus improving the efficiency of NAS. Low-fidelity estimation methods accelerate NAS by using the validation accuracy obtained after training architectures for fewer epochs (namely early-stopping) \cite{li2016hyperband,falkner2018bohb, zoph2018learning, zela2018towards}, training a down-scaled model with fewer cells during the search phase \cite{zoph2018learning, real2019regularized}, or training on a subset of the data \cite{klein2016fast}. However, low-fidelity estimates underestimate the true performance of the architecture and can change the relative ranking among architectures \cite{elsken2018neural}. This undesirable effect on relative ranking is more prominent when the cheap approximation set-up is too dissimilar to the full training \cite{zela2018towards}. As shown in Fig. \ref{fig:baseline_compare} below, the validation accuracy at early epochs of training suffers low rank correlation with the final test performance.  Another class of performance estimation methods trains a regression model to extrapolate the learning curve from what is observed in the initial phase of training. Regression model choices that have been explored include Gaussian processes with a tailored kernel function \cite{domhan2015speeding}, an ensemble of parametric functions \cite{domhan2015speeding}, a Bayesian neural network \cite{klein2016learning} and more recently a $\nu$-support vector machine regressor ($\nu$-SVR)\cite{baker2017accelerating} which achieves state-of-the-art prediction performance \cite{white2021powerful}. Although these model-based methods can often predict the performance ranking better than their model-free early-stopping counterparts, they require a relatively large amount of fully evaluated architecture data (e.g. $100$ fully evaluated architectures in \cite{baker2017accelerating}) to train the regression surrogate properly and optimise the model hyperparameters in order to achieve a good prediction performance. The high computational cost of collecting the training set makes such model-based methods less favourable for NAS unless the practitioner has already evaluated hundreds of architectures on the target task. Moreover, both low-fidelity estimates and learning curve extrapolation estimators are empirically developed and lack theoretical motivation.

Weight sharing is employed in one-shot or gradient-based NAS methods to reduce computational costs \cite{Pham2018_ENAS,Liu2019_DARTS,Xie19_SNAS}. 
Under the weight-sharing setting, all architectures are considered as subnetworks of a supernetwork. Only the weights of the supernetwork are trained while the architectures (subnetworks) inherit the corresponding weights from the supernetwork. This removes the need for retraining each architecture during the search and thus achieves a significant speed-up. However, the weight sharing ranking among architectures often correlates poorly with the true performance ranking \cite{Yang2020NASEFH,Yu2020Evaluating, Zela2020NAS-Bench-1Shot1:}, meaning architectures chosen by one-shot NAS are likely to be sub-optimal when evaluated independently \cite{Zela2020NAS-Bench-1Shot1:}. In Section \ref{subsec:ws_nas}, we demonstrate that we improve the performance of weight sharing in correctly ranking architectures by combining our estimator with it.

Recently, several works propose to estimate network performance without training by using methods from the pruning literature \cite{abdelfattah2021zerocost} or examining the covariance of input gradients across different input images \cite{mellor2020neural}. Such methods incur near-zero computational costs but their performances are often not competitive and do not generalise well to larger search spaces, as shown in Section \ref{subsec:comparison} below. Moreover, these methods can not be improved with additional training budget.


\section{Experiments} \label{sec:experiments}

In this section, we first evaluate the quality of our proposed estimators in predicting the generalisation performance of architectures against a number of baselines (Section \ref{subsec:comparison}), and then demonstrate that simple incorporation of our estimators can significantly improve the search speed and/or quality of both query-based and weight-sharing NAS (Sections \ref{subsec:query_nas} and \ref{subsec:ws_nas}).

We measure the true generalisation performance of architectures with their final test accuracy after being completely trained for $T_{end}$ epochs. To ensure fair assessment of the architecture performance only, we adopt the common NAS protocol where all architectures searched/compared are trained and evaluated under the \emph{same} set of hyper-parameters. Also following \cite{ying2019bench} and \cite{Dong2020nasbench201}, we compare different estimators based on their Spearman's rank correlation which measures how well their predicted ranking correlates with the true test ranking among architectures. 

We compare the following performance estimation methods: our proposed estimators \textbf{TSE}, \textbf{TSE-EMA} and \textbf{TSE-E} described in Section \ref{sec:tse} and simply \textbf{the training losses at each mini batch (TLmini)}. \textbf{Sum of validation losses over all preceding epochs (SoVL)}\footnote{Note, we flip the sign of TSE/TSE-EMA/TSE-E/SoVL/TLmini (which we want to minimise) to compare to the Spearman's rank correlation of the other methods (which we want to maximise).} is similar to TSE but uses the \emph{validation} losses. \textbf{Validation accuracy at an early epoch (VAccES)} corresponds to the early-stopping practice whereby the user estimates the final test performance of a network using its validation accuracy at an early epoch $T < T_{end}$. \textbf{Learning curve extrapolation (LcSVR)} method is the state-of-the-art extrapolation method proposed in \cite{baker2017accelerating} which trains a $\nu$-SVR on previously evaluated architecture data to predict the final test accuracy of new architectures. The inputs for the SVR regression model comprise architecture meta-features and learning curve features up to epoch $T$. In our experiments, we 
optimise the SVR hyperparameters via cross-validation following  \cite{baker2017accelerating}. Three recently proposed zero-cost baselines are also included: an estimator based on \textbf{input Jacobian covariance (JavCov)} \cite{mellor2020neural} and two adapted from pruning techniques \textbf{SNIP} and \textbf{SynFlow} \cite{abdelfattah2021zerocost}. 

We run experiments on architectures generated from a diverse set of NAS search spaces listed in Table \ref{tab:search_spaces} to show that our estimators generalise well (more details are provided in Appendix B). Note $N_{samples}=6466$ for NASBench-201 (NB201) as it's the number of unique architectures in the space. We use the architecture information released in NAS-Bench-301 \cite{siems2020bench} for DARTS and in  \cite{Radosavovic2019} for ResNet and ResNeXt. As for RandWiredNN (RWNN) search space \cite{xie2019exploring,ru2020}, although the number of possible randomly wired architectures are immense, they are generated via a random graph generator which is defined by $3$ hyperparameters. We thus uniformly sampled $69$ sets of hyperparameter values for the generator and generated $8$ randomly wired neural networks from each hyperparameter value, leading to $N_{samples} = 69 \times 8 = 552$. Due to space constraints, we include the results on selecting among the generator hyperparameters for RWNN in Appendix E. All the experiments were conducted on an internal cluster of 16 RTX2080 GPUs.

\begin{table}[t]
\vspace{-0.5cm}
\caption{NAS search spaces used. The true test accuracy of architectures from each search space is obtained after training with SGD on the corresponding image datasets for $T_{end}$ epochs.  $N_{total}$ denotes the total possible architectures exist in the search space and $N_{samples}$ denotes the number of architectures we sample/generate for our experiments.}
\label{tab:search_spaces}\resizebox{1.0\linewidth}{!}{
\begin{tabular}{@{}lllll@{}}
\toprule
Search space & $T_{end}$ & $N_{samples}$ & $N_{T} $                  & Image datasets                     \\ \midrule
NASBench-201 (NB201) \cite{Dong2020nasbench201}  & 200       & 6466       & 15625                      & CIFAR10, CIFAR100, ImageNet-16-120 \\
DARTS   \cite{Liu2019_DARTS, siems2020bench}      	 & 100       & 5000       & $\mathcal{O} (2^{42})$  & CIFAR10                            \\ 
ResNet/ResNeXt   \cite{Radosavovic2019}     & 100      & 50000  & $\mathcal{O} (2^{26})$ & CIFAR10  \\ 
RandWiredNN (RWNN) \cite{xie2019exploring, ru2020}  & 250       & 69 $\times$ 8     & $\mathcal{O} (2^{378})$ & Flower102                         \\
\bottomrule
\end{tabular}} \vspace{-0.5cm}
\end{table}

\subsection{Hyperparameter of TSE estimators} \label{subsec:hyper_sotl}

Our proposed TSE estimators require very few hyperparameters: the summation window size $E$ for TSE-E and the decay rate $\gamma$ for TSE-EMA, and we show empirically that our estimators are robust to these hyperparameters.
For the former, we test different summation window sizes on various search spaces and image datasets in Appendix D and find that $E=1$ consistently gives the best results across all cases. This, together with the almost monotonic improvement of our estimator's rank correlation score over the training budgets, supports our hypothesis discussed in Section \ref{sec:tse} that training information in the more recent epochs is more valuable for performance estimation. Note that TSE-E with $E=1$ corresponds to the sum of training losses over all the batches in one single epoch. As for $\gamma$, we show in Appendix D that TSE-EMA is robust to a range of popular choices $\gamma \in [0.9, 0.95, 0.99, 0.999]$ across various datasets and search spaces. Specifically, the performance difference among these $\gamma$ values are almost indistinguishable compared to the difference between TSE-EMA and TSE-E. Thus, we set $E=1$ and $\gamma=0.999$ in all the following experiments and recommend them as the default choice for potential users who want to apply TSE-E and TSE-EMA on a new task without additional tuning.

\subsection{Comparison of Performance Estimation Quality} \label{subsec:comparison}

\begin{figure}[t]
     \centering
    \begin{subfigure}{0.24\linewidth}
     \centering
    \includegraphics[trim=0.2cm 0.2cm 0.3cm  0.65cm, clip, width=1.0\linewidth]{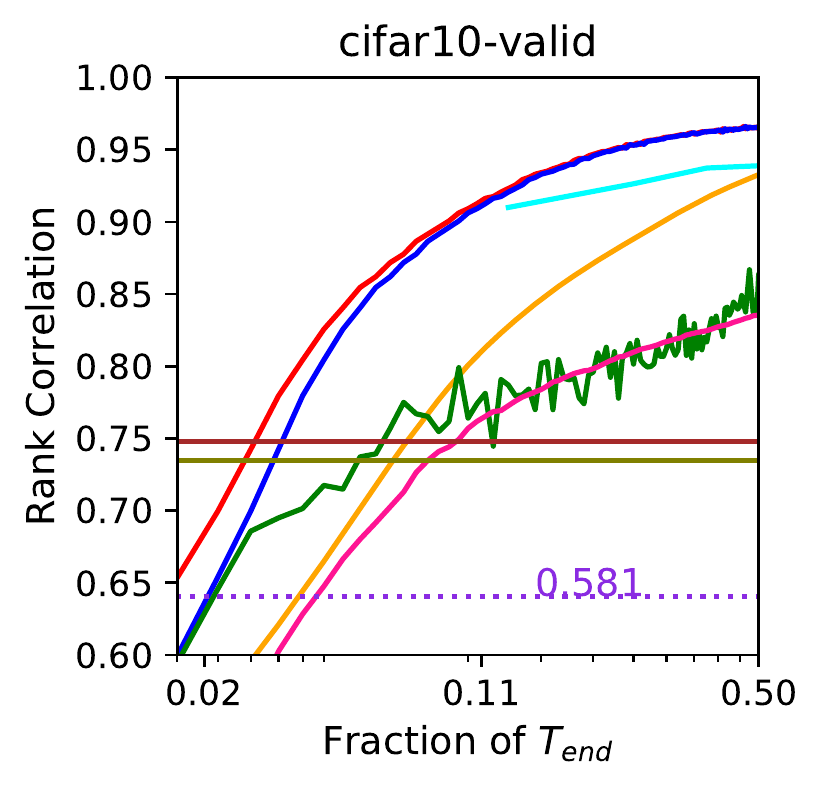}
    \caption{NB201-CIFAR10}
    \end{subfigure}
    \begin{subfigure}{0.24\linewidth}
     \centering
    \includegraphics[trim=0.2cm 0.2cm 0.3cm  0.65cm, clip, width=1.0\linewidth]{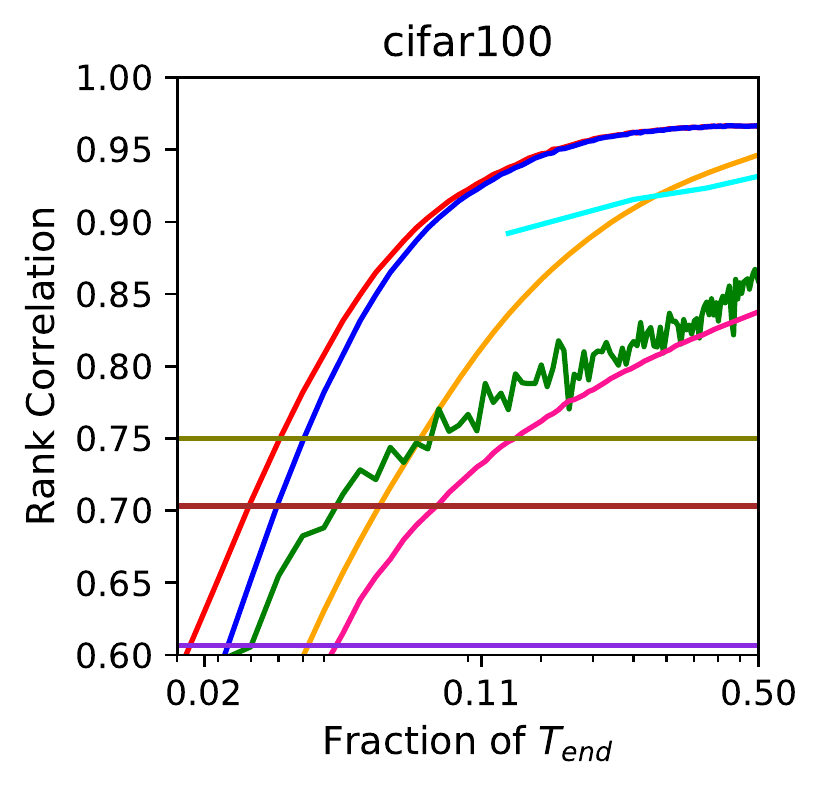}
    \caption{NB201-CIFAR100}
    \end{subfigure}
    \begin{subfigure}{0.24\linewidth}
     \centering
    \includegraphics[trim=0.2cm 0.2cm 0.3cm  0.66cm, clip, width=1.0\linewidth]{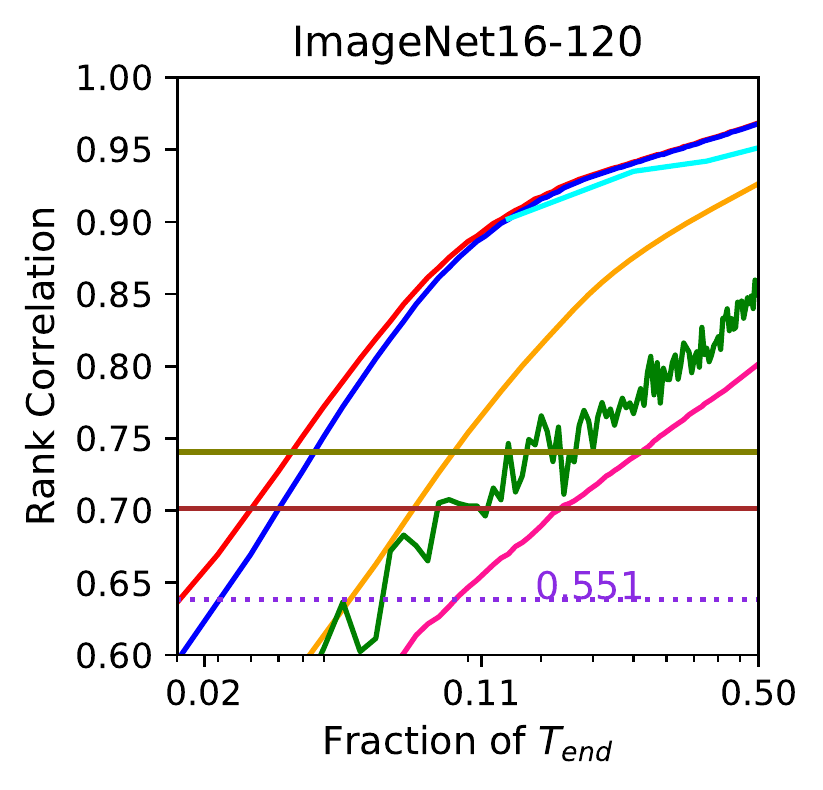}
    \caption{NB201-ImageNet}
    \end{subfigure}
    \begin{subfigure}{0.24\linewidth}
     \centering
    \includegraphics[trim=0.2cm 0.2cm 0.3cm  0.66cm, clip, width=1.0\linewidth]{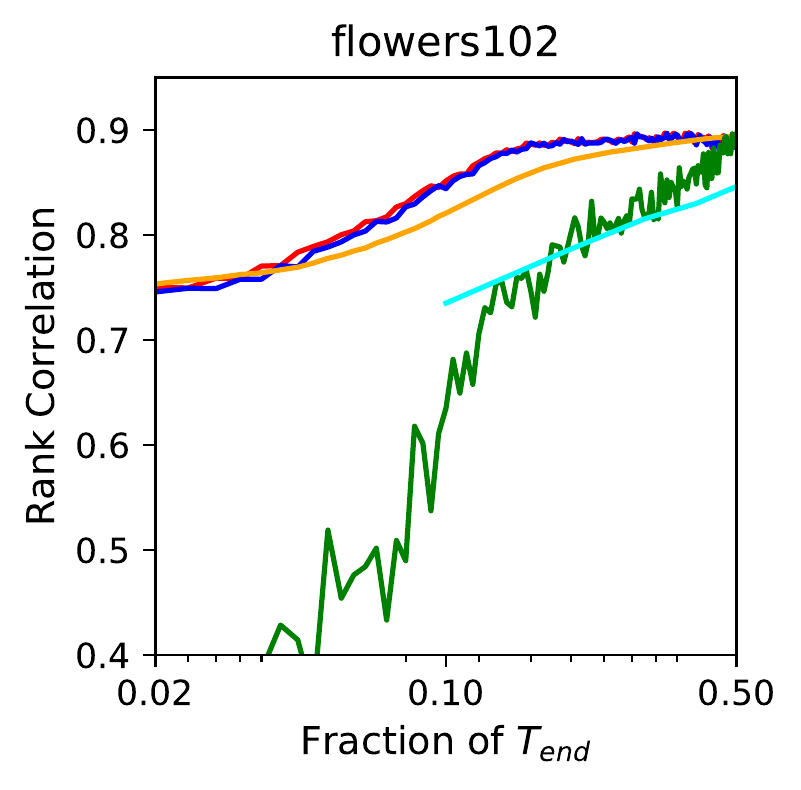}
    \caption{RWNN-Flower102}
    \end{subfigure}
      
       \begin{subfigure}{0.24\linewidth}
     \centering
    \includegraphics[trim=0.2cm 0.2cm 0.3cm  0.65cm, clip, width=1.0\linewidth]{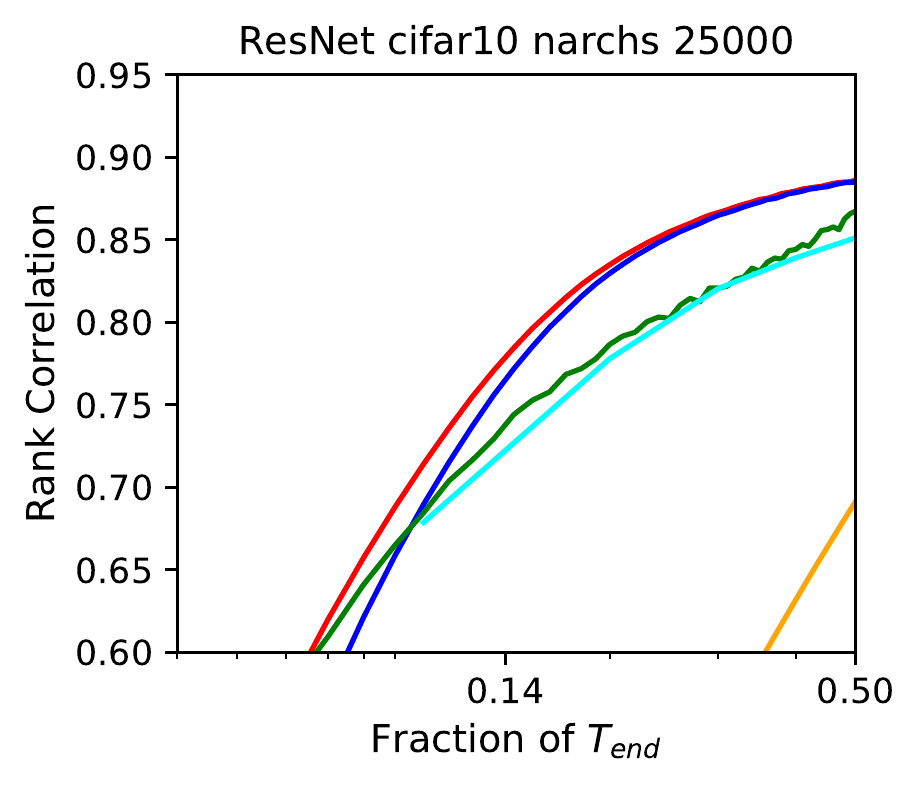}
    \caption{ResNet all}
    \end{subfigure}
    \begin{subfigure}{0.24\linewidth}
     \centering
    \includegraphics[trim=0.2cm 0.2cm 0.3cm  0.65cm, clip, width=1.0\linewidth]{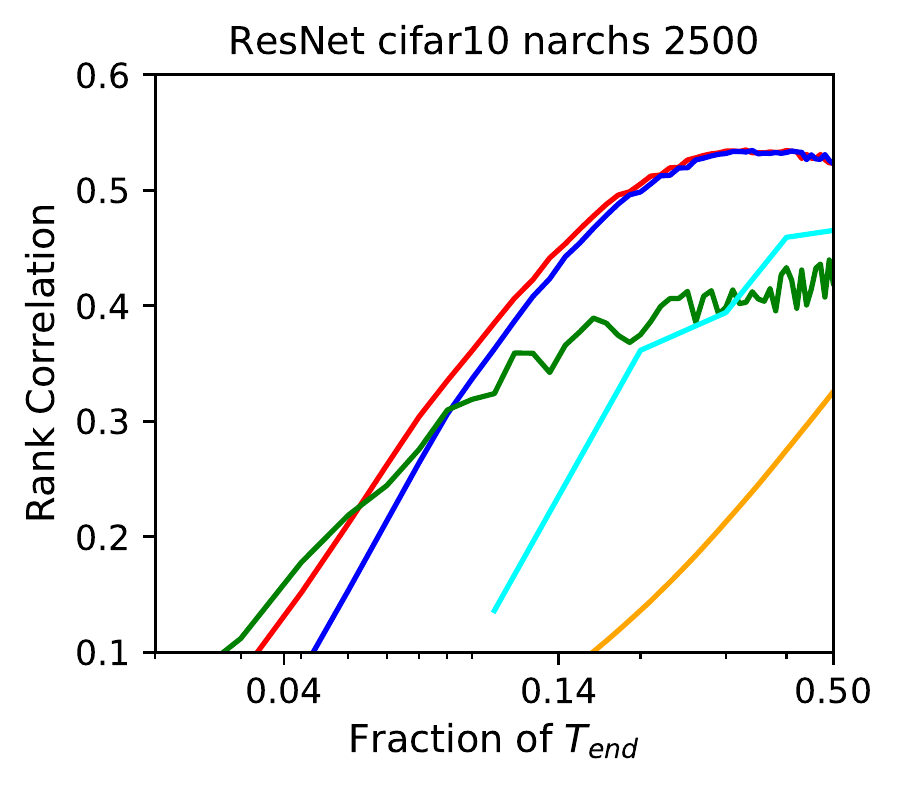}
    \caption{ResNet top $1\%$}
    \end{subfigure}
    \begin{subfigure}{0.24\linewidth}
     \centering
    \includegraphics[trim=0.2cm 0.2cm 0.3cm  0.66cm, clip, width=1.0\linewidth]{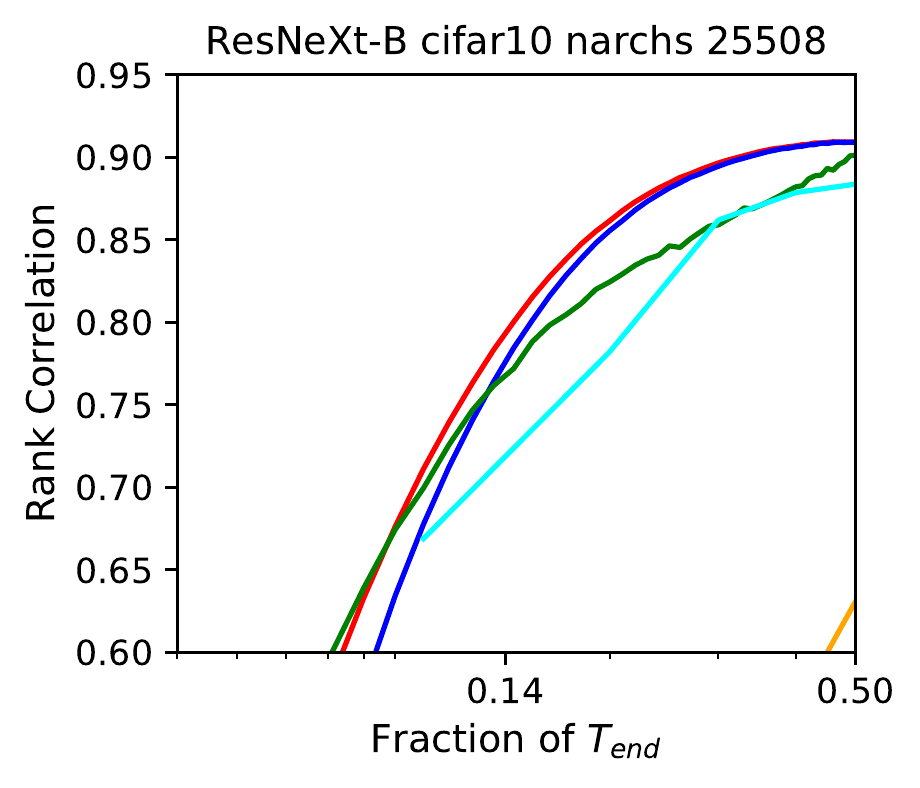}
    \caption{ResNeXt all}
    \end{subfigure}
      \begin{subfigure}{0.24\linewidth}
     \centering
    \includegraphics[trim=0.2cm 0.2cm 0.3cm  0.66cm, clip, width=1.0\linewidth]{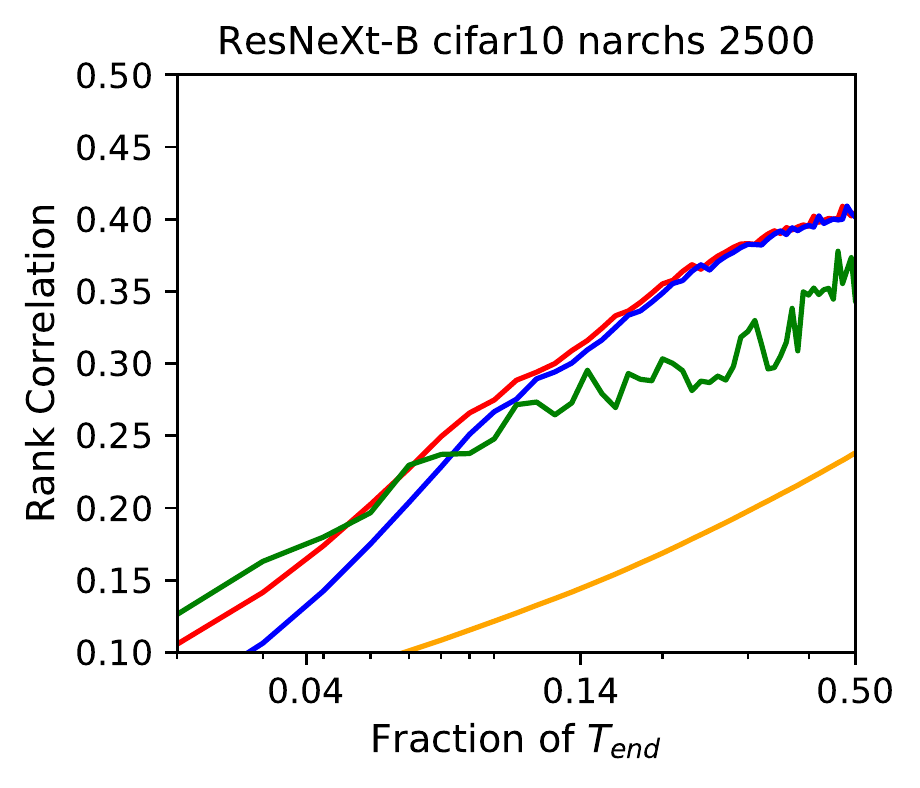}
    \caption{ResNeXt top $1\%$}
    \end{subfigure}
    
        \begin{subfigure}{1.0\linewidth}
    \includegraphics[trim=0cm 0.0cm 0cm  0cm, clip, width=1.0\linewidth]{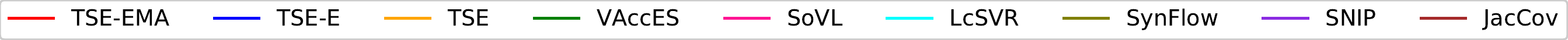} 
      \end{subfigure}
    \caption{Rank correlation performance of various baselines for architectures from a variety of search spaces: (a) to (c) NB201 architectures on three image datasets, (d) RWNNs on Flowers102 and (e) to (h) ResNet and ResNeXt architectures on CIFAR10. In all cases, our TSE-EMA and TSE-E achieve superior rank correlation with the true test performance in much fewer epochs than other baselines. In (f) and (g), we evaluate estimators on the top $1 \%$ of the ResNet/ResNeXt architectures and show that our TSE-EMA and TSE-E can remain competitive on ranking among top architectures, which are particularly desirable for NAS. In (a) and (c), we mark SNIP in a violet dotted line labelled with its rank correlation value as it falls out of the plotted range.} \label{fig:baseline_compare}
    \vspace{-0.2cm}
\end{figure}

\begin{figure}[t]
     \centering
    \begin{subfigure}{0.24\linewidth}
     \centering
    \includegraphics[trim=0.35cm 0.2cm 0.3cm  0.7cm, clip, width=0.95\linewidth]{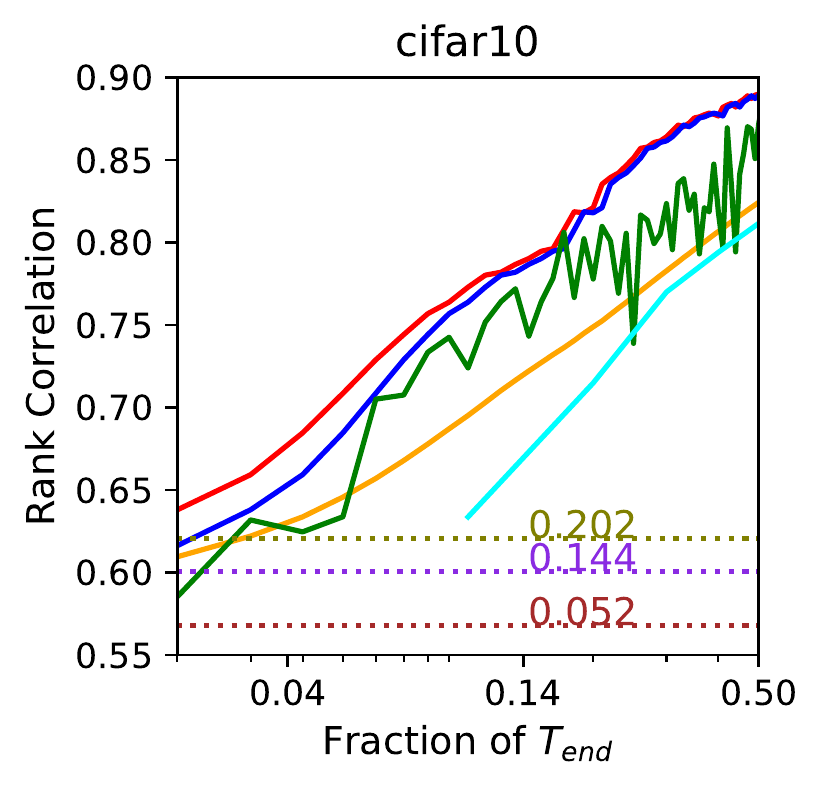}
    \caption{8 cells, lr=0.025, \\ bs=96, cosine scheduler}
    \end{subfigure}
    \begin{subfigure}{0.24\linewidth}
     \centering
    \includegraphics[trim=0.2cm 0.2cm 0.3cm  0.7cm, clip, width=1.0\linewidth]{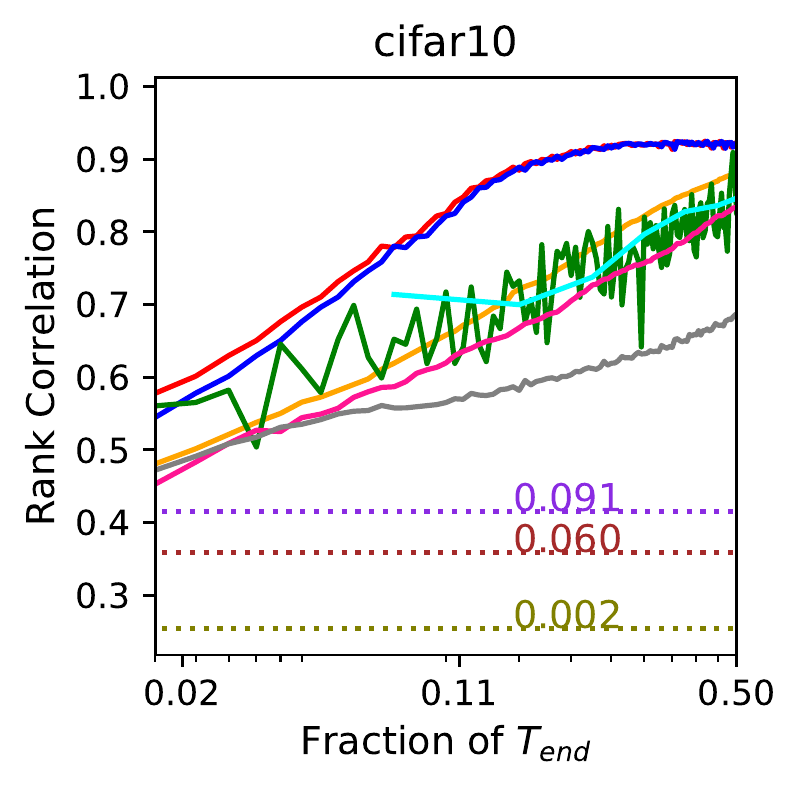}
    \caption{20 cells, lr=0.025, \\bs=96, cosine scheduler}
    \end{subfigure}
        \begin{subfigure}{0.24\linewidth}
     \centering
    \includegraphics[trim=0.2cm 0.2cm 0.3cm  0.7cm, clip, width=1.0\linewidth]{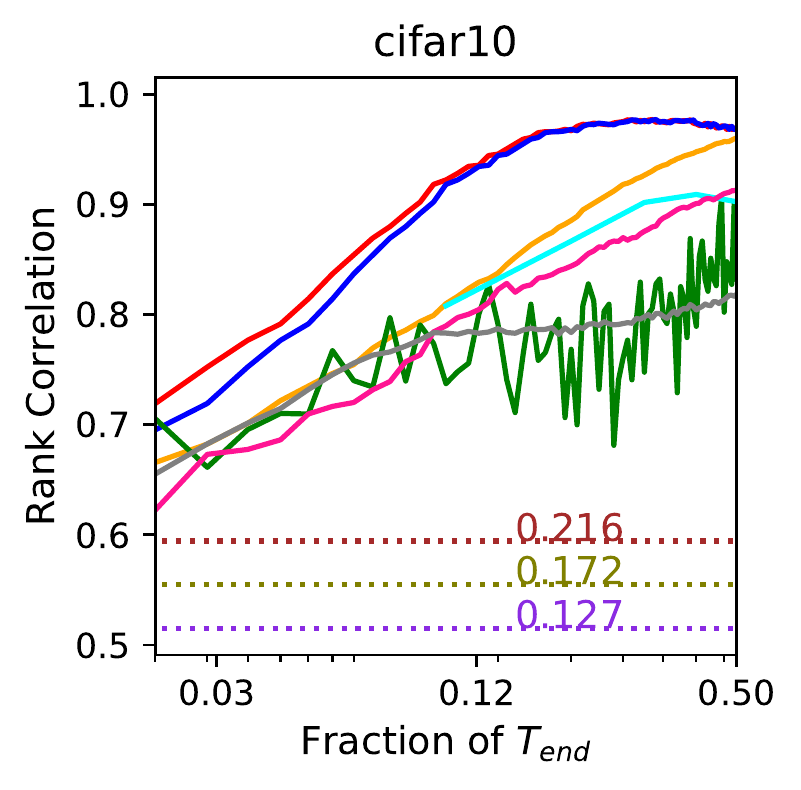}
    \caption{20 cells, lr=0.1, bs=128, step-decay scheduler}
    \end{subfigure}
    \begin{subfigure}{0.24\linewidth}
     \centering
    \includegraphics[trim=0.2cm 0.2cm 0.3cm  0.7cm, clip, width=1.0\linewidth]{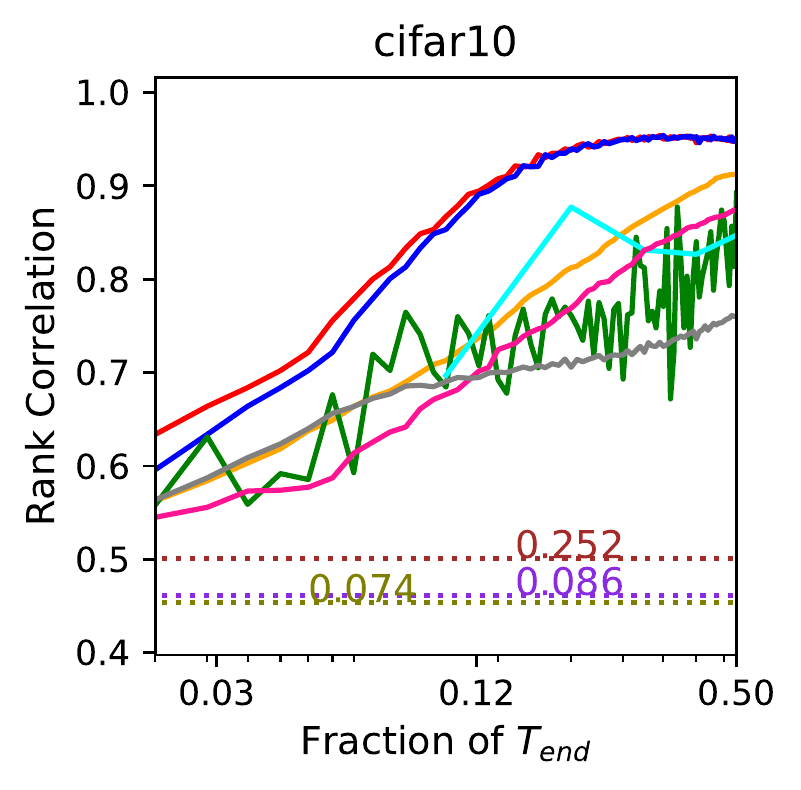}
    \caption{20 cells, lr=0.05, \\ bs=128, cosine scheduler}
    \end{subfigure}
    \begin{subfigure}{1.0\linewidth}
    \includegraphics[trim=0cm 0.0cm 0cm  0cm, clip, width=1.0\linewidth]{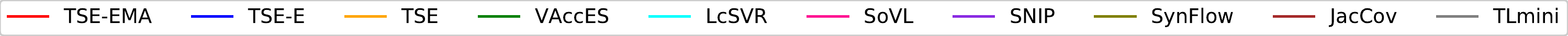}       \vspace{-0.5cm}
    \end{subfigure}
    \caption{Rank correlation performance of various baselines for $5000$ small 8-cell architectures (a) and $150$ large $20$-cell architectures (b) to (d) from DARTS search space on CIFAR10. We use NAS-Bench-301 dataset(NAS301) for computing (a) and for large architectures, we test three training hyperparameter set-ups with different initial learning rates, learning rate schedulers and batch sizes as denoted in the subcaptions. On all four settings, our TSE-E again consistently achieves superior rank correlation in fewer epochs than other baselines. Note all three zero-cost estimators perform poorly (below the plotted range) on DARTS search space across all settings. We denote them in dotted lines with their rank correlation value labelled.} \label{fig:baseline_compare_darts}
    \vspace{-0.3cm}
\end{figure}

\paragraph{Robustness across different NAS search spaces} We now compare our TSE estimators against a variety of other baselines. To mimic the realistic NAS setting \cite{elsken2018neural}, we assume that all the estimators can only use the information from early training epochs and limit the maximum budget to $T \leq 0.5 T_{end}$ in this set of experiments. This is because NAS methods often need to evaluate hundreds of architectures or more during the search \cite{ru2020} and thus rarely use evaluation budget beyond $0.5 T_{end}$ so as to keep the search cost practical/affordable. The results on a variety of the search spaces are shown in Fig. \ref{fig:baseline_compare}. Our proposed estimator TSE-EMA and TSE-E, despite their simple form and cheap computation, outperform all other methods under limited evaluation budget $T <  0.5 T_{end} $ for all search spaces and image datasets.  They also remain very competitive on ranking among the top $1\%$ ResNet/ResNeXt architectures as shown in Fig. \ref{fig:baseline_compare}(f) and (g). TSE-EMA achieves superior performance over TSE-E especially when $T$ is small. This suggests that although the training dynamics at early epochs might be noisy, they still carry some useful information for explaining the generalisation performance of the network. The learning curve extrapolation method, LcSVR, is competitive. However, the method requires $100$ fully trained architecture data to fit the regression surrogate and optimise its hyperparamters via cross validation; a large amount of computational resources are needed to collect these training data in practice. The zero-cost measures JacCov and SynFlow achieve good rank correlation at initialisation but is quickly overtaken by TSE-EMA and TSE-E once the training budget exceeds 6-7 epochs. SNIP performs poorly and falls out of the plot range in Fig. \ref{fig:baseline_compare} (a) and \ref{fig:baseline_compare} (c). 

We further validate on the architectures from the more popular search space used in DARTS. One potential concern is that if models are trained using different hyperparameters that influence the learning curve (e.g. learning rate), the prediction performance of our proposed estimators will be affected. However, this is not a problem in NAS because almost all existing NAS methods \cite{Dong2020nasbench201, ying2019bench, xie2019exploring, Liu2019_DARTS, siems2020bench,white2021powerful} search for the optimal architecture under a fixed set of training hyperparameters. We also follow this \emph{fixed-hyperparamter} set-up in our work. Verifying the quality of various estimators for predicting the generalisation performance across \emph{different hyperparameters} lies outside the scope of our work but would be interesting for future work.  

\paragraph{Robustness across different NAS set-ups} Here, we conduct experiments to verify the robustness of our estimators across different NAS set-ups. On top of the architecture data from NAS-Bench-301 \cite{siems2020bench}, we also generate several additional architecture datasets; each dataset correspond to a different set-up (e.g. different architecture depth, initial learning rate, learning rate scheduler and batch size) and contains $150$ large 20-cell architectures which are randomly sampled from the DARTS space and evaluated on CIFAR10.  The results in Fig. \ref{fig:baseline_compare_darts} show that our estimator consistently outperforms all the competing methods in comparing architectures under different NAS set-ups. Note here the curve of \textbf{TLmini} corresponds to the \emph{average} rank correlation between final test accuracy and the mini-batch training loss over the epoch. The clear performance gain of our TSE estimators over TLmini supports our claim that it is the sum of training losses, which measures the training speed and thus carries the theoretical motivations explained in Section \ref{sec:tse}, instead of simply the training loss at a single minibatch, that gives a good estimation of generalisation performance. Note the rank correlation of all zero-cost measures drop significantly (e.g. SynFlow drops from $0.74$ on NB201 to below $0.2$) on the DARTS search space, and even do worse than the training losses at the first few minibatches (TLmini at T=1). Such inconsistent prediction performance, especially doing weakly on the more practical search space, is undesirable for real-world NAS applications.

\begin{figure}[t]
         \centering
	\begin{subfigure}{0.45\linewidth}
     \centering
    \includegraphics[trim=0.25cm 0.3cm 0.2cm  0.2cm, clip, width=0.8\linewidth]{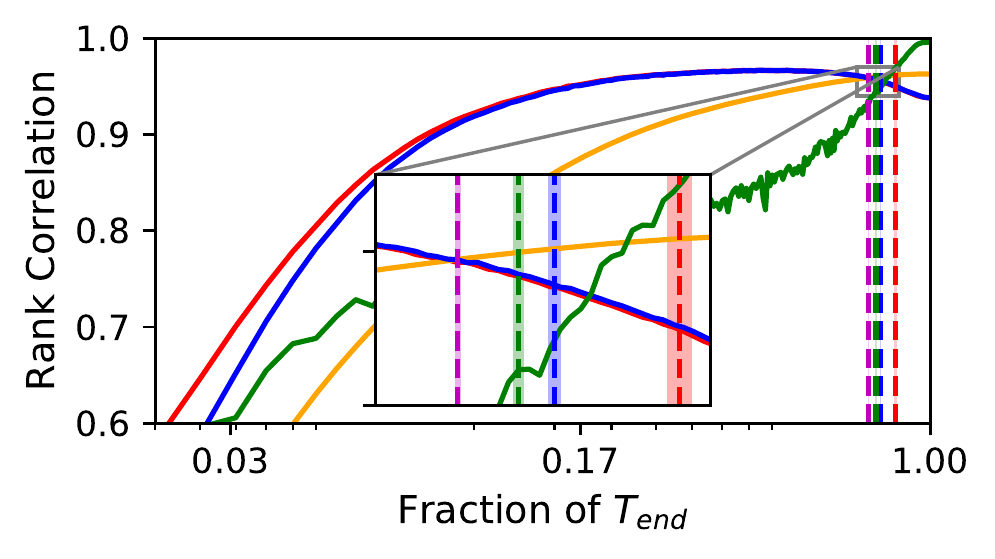}
    \caption{NB201-CIFAR10}
    \end{subfigure}
    \hspace{0.5cm}
	\begin{subfigure}{0.45\linewidth}
     \centering
    \includegraphics[trim=0.25cm 0.3cm 0.2cm  0.2cm, clip, width=0.8\linewidth]{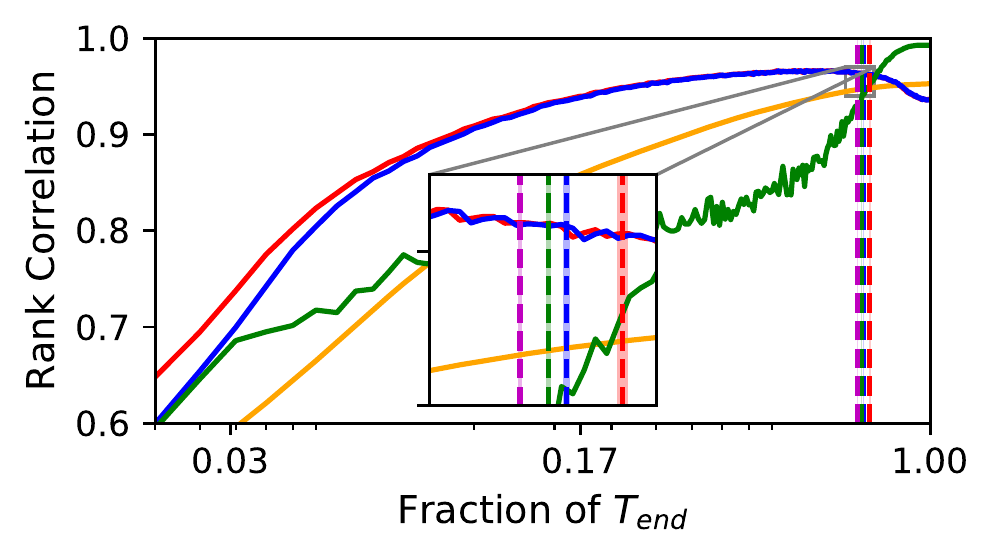}
    \caption{NB201-CIFAR100}
    \end{subfigure}
         \begin{subfigure}{1.0\linewidth}
    \includegraphics[trim=0cm 0.0cm 0cm  0cm, clip, width=1.0\linewidth]{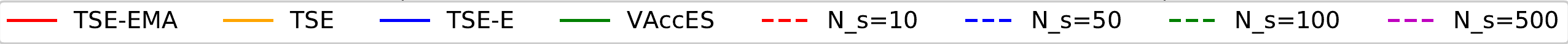} \vspace{-0.5cm}
    \end{subfigure}
    \caption{Rank correlation performance up to $T= T_{end}$. If the users want to apply our estimators for large training budget, they can estimate the effective range of our estimators based on the minimum epoch $T_{o}$ when overfitting happens among the $N_s$ observed architectures. They can then stop our estimators early at $0.9T_{o}$(marked by vertical lines) or switch back to validation accuracy beyond that. }\label{fig:termination_criterion}\vspace{-0.5cm}

\end{figure}

\paragraph{Procedure to decide the effective training budget}  We include two examples showing the performance of our estimator for training budgets beyond $0.5 T_{end}$ in Fig. \ref{fig:termination_criterion}. Our estimators can remain superior for a relatively wide range of training budgets. Although they will eventually be overtaken by validation accuracy as the training budget approaches $T_{end}$ as discussed in Section \ref{sec:tse},  the region requiring large training budget is less interesting for NAS where we want to maximise the cost-saving by using performance estimators. However, if the user wants to apply our estimators with a relatively large training budget, we propose a simple method here to estimate when our estimators would be less effective than validation accuracy. We notice that that our estimators, TSE-EMA and TSE-E, become less effective when the architectures compared start to overfit because both of them rely heavily on the lastest-epoch training losses to measure training speed, which is difficult to estimate when the training losses become too small. 
Thus, if we observe one of the architectures compared overfits beyond $T_{o} < T_{end}$, we can stop the computation of TSE-E and TSE-EMA early by reverting to a checkpoint at $T=0.9 T_{o}$. We randomly sample $N_s=10, 50, 100, 500$ architectures out of all architectures compared and assume that we have access to their full learning curves. We then decide the threshold training budget $0.9 T_{o}$ (vertical lines) as the minimum training epoch that overfitting happens among these $N_s$ architectures. We repeat this for 100 random seeds and plot the mean and standard error of the threshold for each $N_s$ in Fig. \ref{fig:termination_criterion}. It's evident that we can find a quite reliable threshold with a sample size as small as $N_s=10$. 
Please refer to Appendix F for more analyses.

%

\subsection{Speed up Query-based NAS} \label{subsec:query_nas}

\begin{figure}[t]
         \centering
	\begin{subfigure}{0.495\linewidth}
     \centering
    \includegraphics[trim=0.25cm 0.3cm 0.3cm  0.3cm, clip, width=0.33\linewidth]{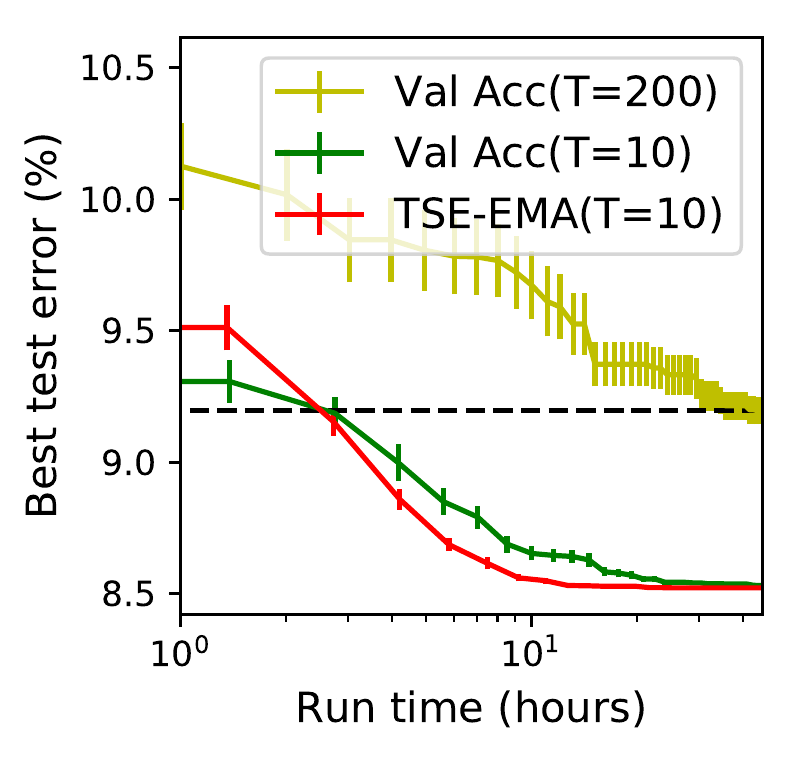}
        \includegraphics[trim=0.25cm 0.3cm 0.3cm  0.3cm, clip, width=0.32\linewidth]{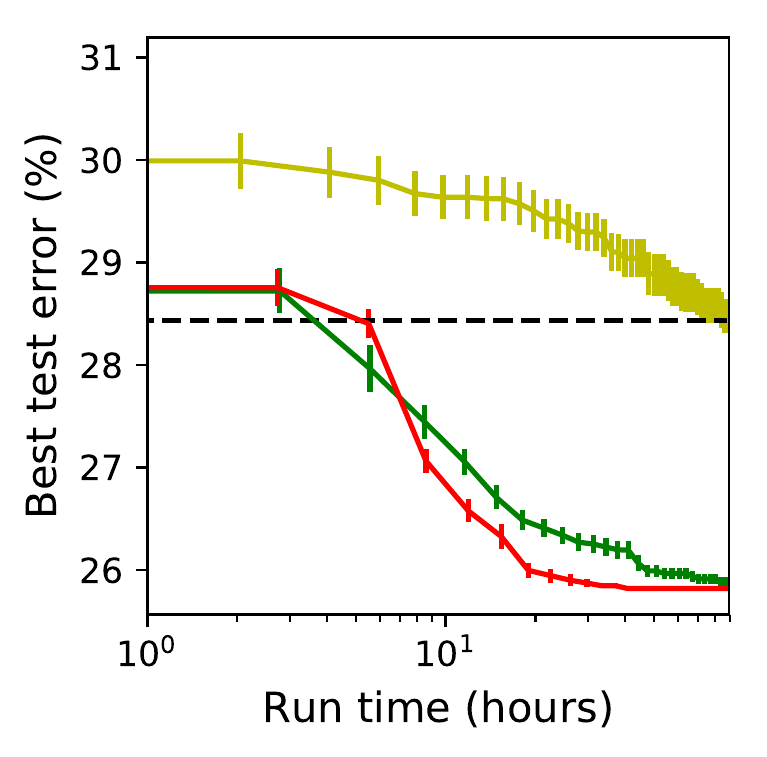}
    \includegraphics[trim=0.25cm 0.3cm 0.3cm  0.3cm, clip, width=0.32\linewidth]{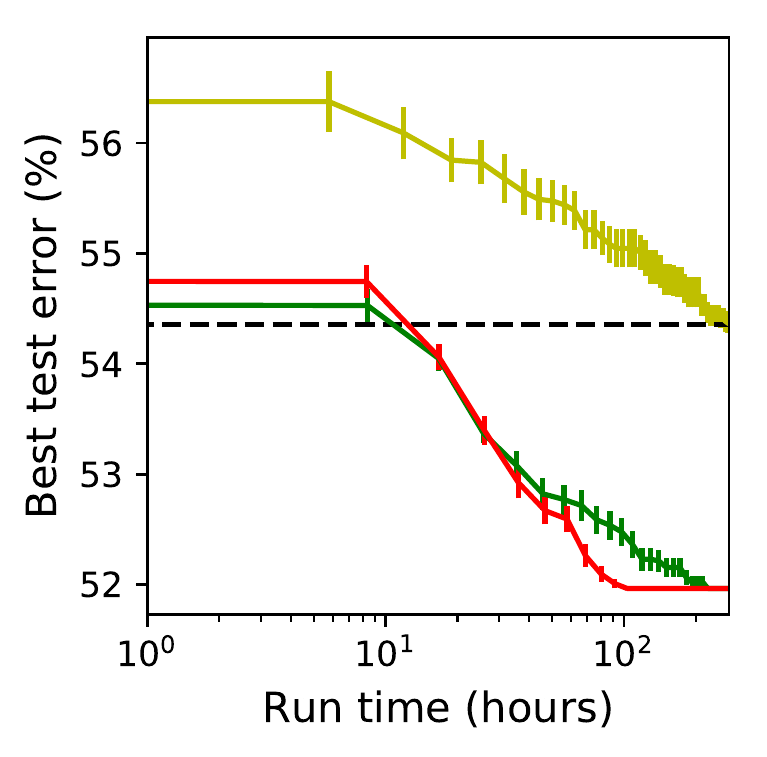}
    \caption{Regularised Evolution (RE) }
    \end{subfigure}
	\begin{subfigure}{0.495\linewidth}
     \centering
    \includegraphics[trim=0.25cm 0.3cm 0.3cm  0.3cm, clip, width=0.33\linewidth]{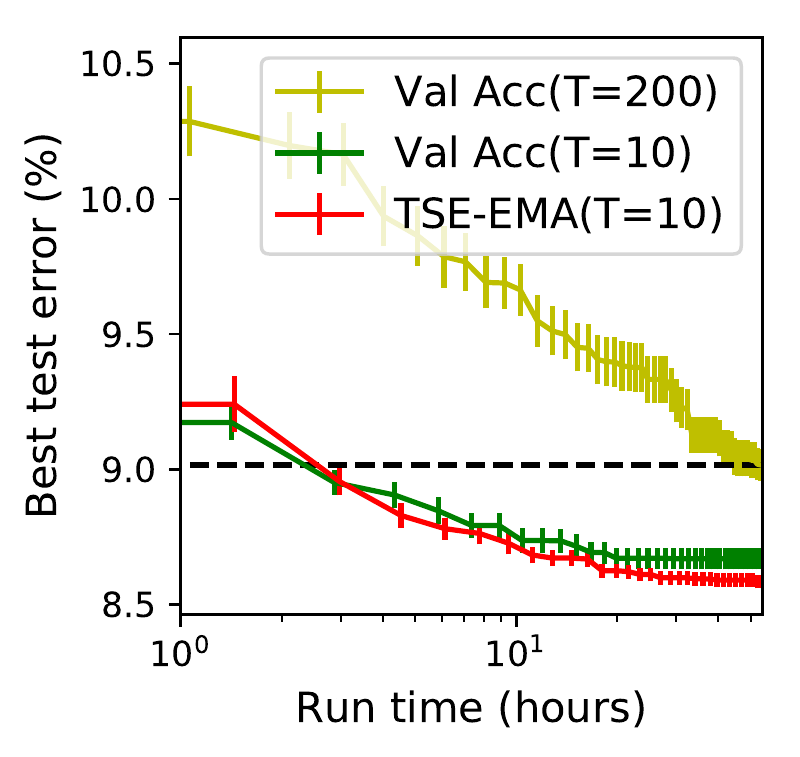}
    \includegraphics[trim=0.25cm 0.3cm 0.3cm  0.3cm, clip, width=0.32\linewidth]{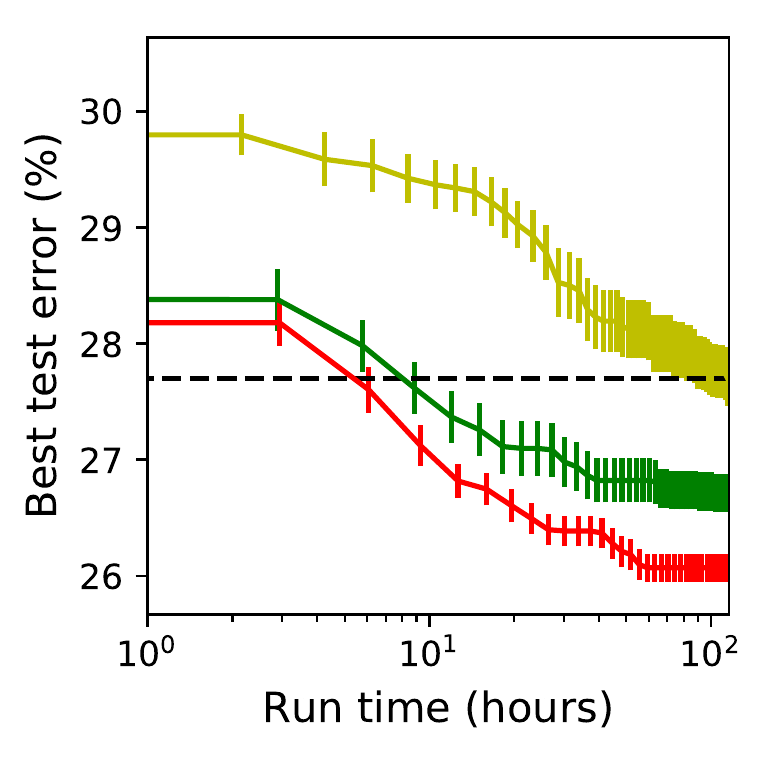}
    \includegraphics[trim=0.25cm 0.3cm 0.3cm  0.3cm, clip, width=0.32\linewidth]{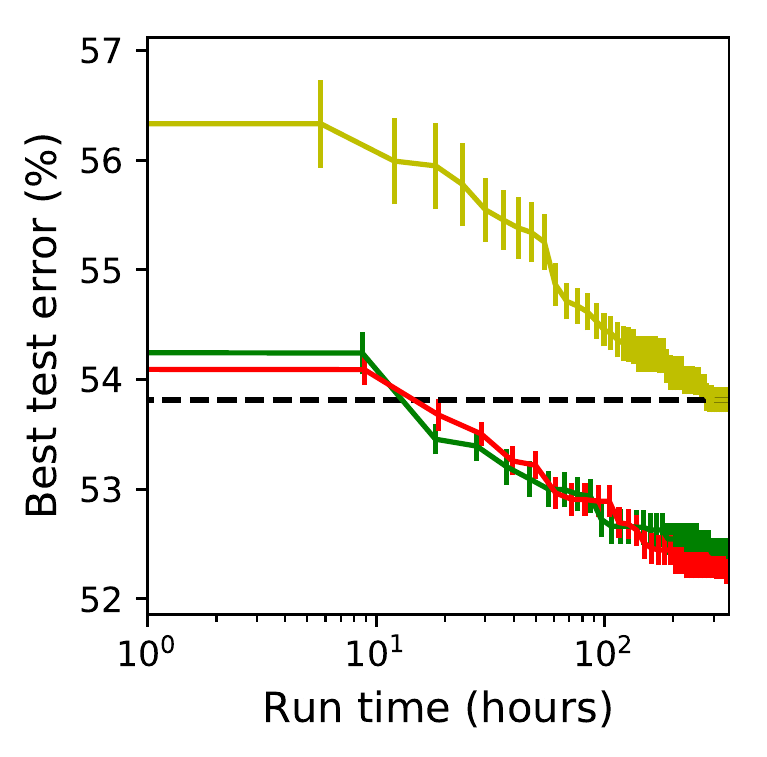}
    \caption{Bayesian Optimisation (BO)}
    \end{subfigure}
    \caption{NAS performance of RE and BO in combined with final validation accuracy Val Acc (T=200), early-stopping validation accuracy Val Acc (T=10) and our estimator TSE-EMA(T=10) on NB201. For each subplot, we experiment on the three image datasets: on CIFAR10 (left), CIFAR100 (middle) and ImageNet (right). TSE-EMA leads to the fastest convergence to the top performing architectures in all cases. The black dashed line is to facilitate the comparison of runtime taken to reach a certain test error among different variants.}\label{fig:retpe_nas}
    \vspace{-0.3cm}
\end{figure}

In this section, we demonstrate the usefulness of our estimator for NAS by incorporating TSE-EMA, at $T=10$ into several query-based NAS search strategies: Regularised Evolution \cite{real2019regularized} (a) in Fig. \ref{fig:retpe_nas}), Bayesian Optimisation \cite{bergstra2011algorithms} (b) in Fig. \ref{fig:retpe_nas}) and Random Search \cite{bergstra2012random} (Appendix G). We perform architecture search on NB201 search space. We compare this against the other two benchmarks which use the final validation accuracy at $T=T_{end}=200$, denoted as Val Acc (T=200) and the early-stop validation accuracy at $T=10$, denoted as Val Acc (T=10), respectively to evaluate the architecture's generalisation performance. All the NAS search strategies start their search from 10 random initial data and are repeated for 20 seeds. 
The mean and standard error results over the search time are shown in Fig. \ref{fig:retpe_nas}. By using our estimator, the NAS search strategies can find architectures with lower test error given the same time budget or identify the top performing architectures using much less runtime as compared to using final or early-stopping validation accuracy. The gain of using our estimator is more significant for NAS methods performing both \emph{exploitation} and exploration (RE and BO) than that doing pure exploration (Random Search in Appendix G). 

\subsection{Improving One-shot and Gradient-based NAS}\label{subsec:ws_nas}

\begin{table}[t]
\caption{Results of performance estimators in one-shot NAS setting over 3 supernetwork training initialisations. For each supernetwork, we randomly sample 500 random subnetworks for DARTS and 200 for NB201, and compute their TSE, Val Acc after inheriting the supernetwork weights and training for $B$ additional minibatches. Rank correlation measures the estimators' correlation with the rankings of the true test accuracies of subnetworks when \emph{trained from scratch independently}, and we compute the average test accuracy of the top 10 architectures identified by different estimators from all the randomly sampled subnetworks.}
\label{tab:one_shot_nas}
    	\vspace{2mm} \resizebox{1.0\linewidth}{!}{
\begin{tabular}{@{}cccccccccc@{}}\toprule
\multirow{3}{*}{B}   & \multirow{3}{*}{Estimator} & \multicolumn{4}{c}{Rank Correlation}                                                     & \multicolumn{4}{c}{Average Accuracy of Top 10 Architectures}                                 \\\cmidrule(lr){3-6}\cmidrule(l){7-10}
                     &                            & \multicolumn{3}{c}{NB201-CIFAR10}                                         & DARTS               & \multicolumn{3}{c}{NB201-CIFAR10}                                      & DARTS                \\\cmidrule(lr){3-5}\cmidrule(l){6-6} \cmidrule(l){7-9}\cmidrule(l){10-10}
                     &                            & RandNAS              & FairNAS              & MultiPaths           & RandNAS             & RandNAS             & FairNAS             & MultiPaths          & RandNAS              \\\midrule
\multirow{2}{*}{100} & TSE                        & \textbf{0.70 (0.02)} & \textbf{0.84 (0.01)} & \textbf{0.83 (0.01)}         & \textbf{0.30(0.04)} & \textbf{92.67 (0.12} & \textbf{92.7 (0.1)} & \textbf{92.63 (0.12)} & \textbf{93.64(0.04)} \\
                     & Val Acc                    & 0.44 (0.15)          & 0.56 (0.17)          & 0.67 (0.05)          & 0.11(0.04)          & 91.47 (0.31)          & 91.73 (0.21)        & 91.77 (0.78)          & 93.20(0.04)          \\
                     \midrule
\multirow{2}{*}{200} & TSE                        & \textbf{0.70 (0.03)} &      \textbf{0.850 (0.01)}       & \textbf{0.83 (0.01)} & \textbf{0.32(0.04)} & \textbf{92.70 (0.00)} &    \textbf{92.77 (0.06)}   & \textbf{92.73 (0.06)} & \textbf{93.55(0.04)} \\
                     & Val Acc                    & 0.41 (0.10)          &          0.56 (0.17)        & 0.53 (0.11)          & 0.09(0.02)          & 91.53 (0.55)          &   92.40 (0.10)      & 92.23 (0.23)          & 93.34(0.02)          \\
                     \midrule
\multirow{2}{*}{300} & TSE                        & \textbf{0.71 (0.03)} &   \textbf{0.851 (0.00)}              & \textbf{0.82 (0.01)}          & \textbf{0.34(0.04)} & \textbf{92.70 (0.00)} &      \textbf{92.77 (0.06)}    & \textbf{92.70 (0.00)} & \textbf{93.65(0.04)} \\
                     & Val Acc                    & 0.44 (0.04)          &        0.62 (0.08)              & 0.59 (0.71)          & 0.06(0.02)          & 91.20 (0.35)          &          92.10 (0.50)          & 91.43 (0.72)          & 93.31(0.02)          \\
                     \bottomrule         
\end{tabular}}\vspace{-0.2cm}
\end{table}

Different from query-based NAS strategies, which evaluate the architectures queried by training them independently from scratch,  another popular class of NAS methods use weight sharing to accelerate the evaluation of the validation performance of architectures (subnetworks) and use this validation information to select architectures or update architecture parameters. Here we demonstrate that our TSE estimator can also be a plug-in replacement for validation accuracy or loss used in such NAS methods to improve their search performance. 

\paragraph{One-shot NAS} We first experiment on a classic one-shot method, RandNAS \cite{li2020random}, which trains a supernetwork by uniform sampling,  
then performs architecture search by randomly sampling subnetworks from the trained supernetwork and comparing them based on their validation accuracy. We follow the RandNAS procedure for the supernetwork training but modify the search phase: for each randomly sampled subnetwork, we train it for $B$ additional mini-batches after inheriting weights from the trained supernetwork to compute our TSE estimator. Note although this introduces some costs, our estimator saves all the costs from evaluation on validation set as it doesn't require validation data.
To ensure fair comparison, we also recompute the validation accuracy of each subnetwork after the additional training. We also experiment with more advanced supernetwork training techniques such as FairNAS \cite{chu2019fairnas} and MultiPaths \cite{yu2019universally} and show that our estimators can be applied on top of them to further improve the rank correlation performance. 

We evaluate the rank correlation performance and average test accuracy of the top-10 architectures recommended by different performance estimators among 500 random subnetworks sampled from the DARTS supernet\footnote{We use NAS-Bench-301 to compute the true test accuracy of each subnetwork when trained independently from scratch.} and 200 random subnetworks from the NB201 supernet. We repeat the experiments on $B=100,200,300$ and over $3$ different supernetworks training seeds. The mean and standard deviation results are shown in Table \ref{tab:one_shot_nas}. It is evident that our TSE leads to $170\%$ to $300\%$ increase in rank correlation performance compared to validation accuracy and achieves higher average test accuracy of the top 10 architectures across all supernetwork training techniques and search spaces. This implies that our estimator can lead to architectures with better generalisation performance under the popular weight-sharing setting. We include the results for more estimators in Appendix G.

\begin{figure}[t]
     \centering
    \begin{subfigure}{0.45\linewidth}
     \centering
    \includegraphics[trim=0.3cm 0.4cm 0.3cm  0.0cm, clip, width=0.49\linewidth]{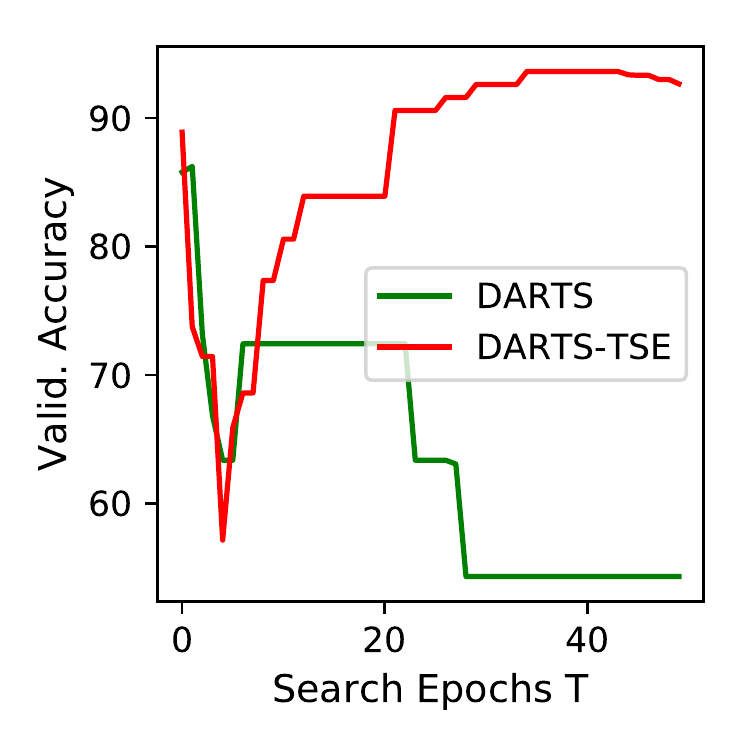}
    \includegraphics[trim=0.3cm 0.4cm 0.3cm  0.0cm, clip, width=0.49\linewidth]{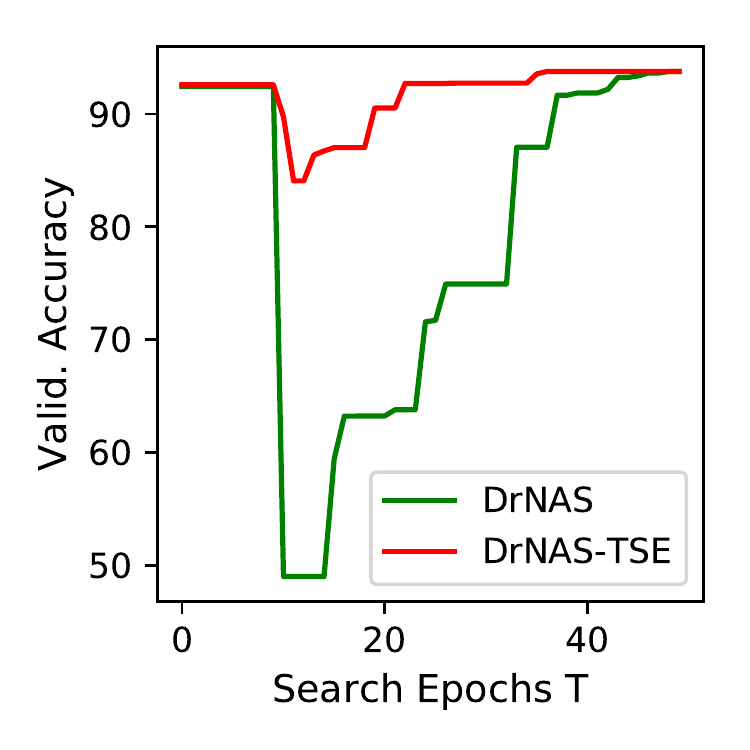}
    \caption{NB201-CIFAR10}
    \end{subfigure}
    \hspace{0.5cm}
    \begin{subfigure}{0.45\linewidth}
     \centering
    \includegraphics[trim=0.3cm 0.4cm 0.3cm  0.0cm, clip, width=0.49\linewidth]{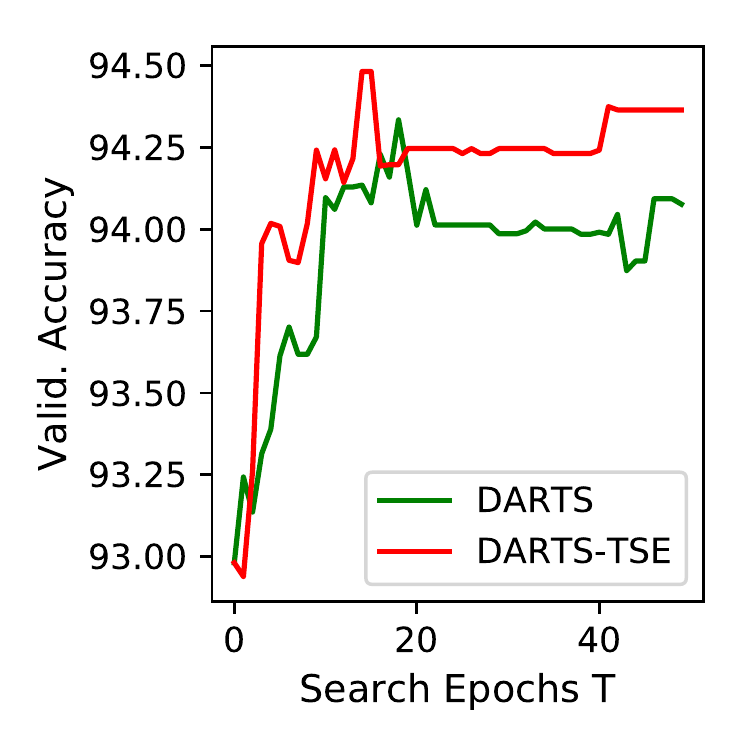}
    \includegraphics[trim=0.3cm 0.4cm 0.3cm  0.0cm, clip, width=0.49\linewidth]{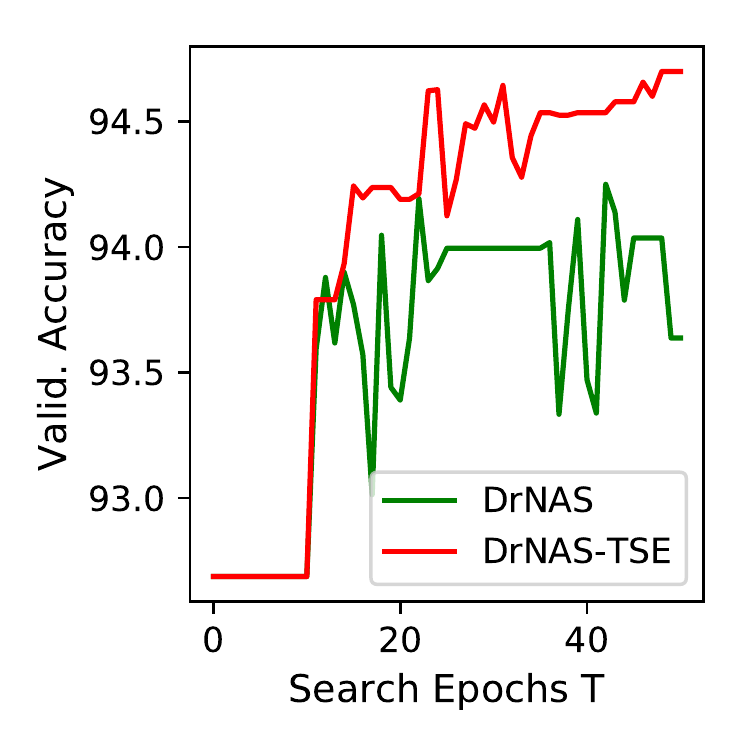}
    \caption{DARTS-NB301}
    \end{subfigure}
   
    \caption{Test accuracy of the subnetwork recommended by differentiable NAS methods over the search epochs. Original DARTS, DrNAS (in green) use the gradient of validation loss to update the architecture parameters but their variants (DARTS-TSE and DrNAS-TSE) (in red) uses that of our estimator computed over 100 mini-batches. TSE help mitigate the overfitting of DARTS on NB201.} \label{fig:darts_drnas}
    \vspace{-3mm}
\end{figure}

\paragraph{Differentiable NAS} 
Finally, we briefly demonstrate the use of our estimators on differentiable NAS. We modify two differentiable approach, DARTS \cite{Liu2019_DARTS} and DrNAS \cite{chen2021drnas} by directly replacing the derivative of the validation loss with that of our TSE estimator computed over 100 mini-batches (B=100 as in one-shot NAS setting above) to update the architecture parameters. Please refer to Appendix G for details of our adaptation. We test both approaches and their TSE variants on the NB201-CIFAR10 as well as DARTS search space. Again we use NAS-Bench-301 to obtain the true test performance of searched DARTS architectures on CIFAR10 (DARTS-NB301). The test accuracy of the subnetwork recommended over the search epochs are shown in Figure \ref{fig:darts_drnas}: we average over 3 seeds for NB201-CIFAR10 and use the default seed for DARTS-NB301. The results show that a simple integration of our estimator into the differentiable NAS frame can lead to clear search performance improvement and even mitigate the overfitting (to skip-connection) problem suffered by DARTS on NB201 search space (a).

\section{Conclusion}
We propose a simple yet reliable method for estimating the generalisation performance of neural architectures based on their training speed measured by the sum of early training losses. Our estimator is theoretically motivated by the connection between training speed and generalisation, and outperforms other efficient estimators in terms of rank correlation with the true test performance under different search spaces as well as different training set-ups. Moreover, it can lead to significant speed-ups and performance gains when applied to different NAS strategies including one-shot and differentiable ones. We believe our estimator can be a very useful tool for achieving efficient neural architecture search. On societal impacts, our estimators, by reducing the computation and time required for performance evaluation during NAS, can save significant amount of carbon emission/environmental costs as AutoML being more widely used in industries, and enable the potential users with limited computation budgets to use NAS.

\appendix
\section*{Appendix}
\section{PAC-Bayesian Generalisation Bounds Estimator} \label{app:pac_bayes}

\begin{corollary}[Corollary 2 of \cite{germain2016pac}]
Given a data distribution $\mathcal{D}$, a model parameter set $\Theta$, a prior distribution $P(\theta)$ over $\Theta$, $\delta \in (0, 1]$, if the negative log likelihood $\ell$ lies within the range $[a, b]$, we have, with probability at least $1-\delta$ over the choice of $(\mathbf{X}, \mathbf{y}) \sim \mathcal{D}^n$, which denotes $n$ input-output pairs sampled from a data distribution,
\begin{equation}
    \mathbb{E}_{\theta \sim \rho} \mathbb{E}_{x, y \sim \mathcal{D}}[\ell(f_\theta(x), y)] \leq a + c\bigg [ 1 - e^a  (Z_{\mathbf{X}, \mathbf{y}}\delta)^{\frac{1}{n} } \bigg ], \label{eq:pacbayescoro2}
\end{equation}
where $\rho$ is the posterior of model parameters $P(\theta | \mathbf{X}, \mathbf{y})$, $Z_{\mathbf{X}, \mathbf{y}}$ is the marginal likelihood and $c=\frac{b - a}{1 -  e^{a-b}}$.
\end{corollary}

\begin{theorem}[Theorem 3.3 of \cite{lyle2020}]
Let $(\mathbf{X}, \mathbf{y}) \sim \mathcal{D}^n$ denote $n$ input-output pairs sampled from a data distribution, $\theta_k$ be generated by Bayesian updates, we can obtain the following lower bound on the log marginal likelihood $\log Z_{\mathbf{X}, \mathbf{y}}$: 
\begin{align}
    \log Z_{\mathbf{X}, \mathbf{y}} &\geq \sum^n_{k=1} \mathbb{E}_{\theta_k \sim P(\cdot \vert \mathbf{X}_{<k}, \mathbf{y}_{<k})} \left[\log P(x_{k}, y_{k} \vert \theta_k) \right] \label{eq:lb_mll}\\
    & \gtrapprox \sum^n_{k=1} \log P(x_{k}, y_{k} \vert \hat{\theta}_k)  \label{eq:mclb_mll}\\
    & = - \sum^n_{k=1} \ell \left(f_{\hat{\theta}_k}(x_k), y_k \right). \label{eq:reformcmlb_mll}
\end{align}
\end{theorem}
Equation \ref{eq:lb_mll} to Equation \ref{eq:mclb_mll} is obtained by using MC approximation to estimate the expectation and Equation \ref{eq:mclb_mll} to Equation \ref{eq:reformcmlb_mll} is because $\ell$ denotes negative log likelihood.

By substituting Equation \ref{eq:reformcmlb_mll} into Equation \ref{eq:pacbayescoro2}, we obtain a PAC-Bayesian bound, which depends on a sum of negative log likelihoods, described in Section 2: 
\begin{align}
    \mathbb{E}_{\theta \sim \rho} & \mathbb{E}_{x, y \sim \mathcal{D}}[\ell(f_\theta(x), y)] \leq a + c\bigg [ 1 - e^a  (Z_{\mathbf{X}, \mathbf{y}}\delta)^{\frac{1}{n} } \bigg ] \nonumber\\
    \approx&  a + c\bigg [ 1 - e^a  \left(e^{- \sum_{k=1}^n \ell \left(f_{\theta}(x_k), y_k \right)}\delta \right)^{\frac{1}{n} } \bigg ]. \label{eq:pac_bayes_est}
\end{align}

\section{Hardware and Datasets Description} \label{app:datasets}
All experiments are performed with an internal cluster of 16 RTX2080 GPUs and a Intel Core i5 CPU. The datasets we experiment with are:
\begin{itemize}
    \item \textbf{NASBench-201} \cite{Dong2020nasbench201}: the dataset contains information of 15,625 different neural architectures, each of which is trained with SGD optimiser and evaluated on 3 different datasets: CIFAR10, CIFAR100, IMAGENET-16-120 for 3 random initialisation seeds. The training accuracy/loss, validation accuracy/loss after every training epoch as well as architecture meta-information such as number of parameters, and FLOPs are all accessible from the dataset. The search space of the NASBench-201 dataset is a 4-node cell and applicable to almost all up-to-date NAS algorithms. The dataset is available at \url{https://github.com/D-X-Y/NAS-Bench-201} with MIT License.
    \item \textbf{RandWiredNN}: we produce this dataset by generating 552 randomly wired neural architectures from the random graph generators proposed in \cite{xie2019exploring} and evaluate their performance on the image dataset FLOWERS102 \cite{nilsback2008automated}. We explore 69 sets of hyperparameter values for the random graph generators and for each set of hyperparameter values, we sample 8 randomly wired neural networks from the generator. A randomly wired neural network comprises 3 cells connected in sequence and each cell is a 32-node random graph. The wiring/connection within the graph is generated with one of the three classic random graph models in graph theory: Erdos-Renyi (ER), Barabasi-Albert (BA) and Watt-Strogatz (WS) models. Each random graph model has 1 or 2 hyperparameters that decide the generative distribution over edge/node connection in the graph.  All the architectures are trained with SGD optimiser for 250 epochs and other training set-ups follow those in \cite{Liu2019_DARTS}. This dataset allows us to evaluate the performance of our simple estimator on hyperparameter/model selection for the random graph generator. We will release this dataset after paper publication. 

    \item \textbf{DARTS}: DARTS search space \cite{Liu2019_DARTS} is more general than those of NASBench-201 and contains over $10^{18}$ architectures. It's also the most widely adopted space in NAS \cite{zoph2018learning, Liu2019_DARTS, Chen2019_PDARTS, Xie19_SNAS, xu2019pc, real2019regularized, li2020random, Pham2018_ENAS, shaw2019meta, zhou2020theory}. Particularly, this search space comprises a cell of 7 nodes: the first two nodes in cell $k$ are the input nodes which equals to the outputs of cell $k-2$ and cell $k-1$ respectively. The last node in the cell $k$ is the output node which gives a depthwise concatenation of all the intermediate nodes. The remaining four intermediate nodes are operation nodes take can take one out of eight operation choices. An architecture from this search space is formed by stacking the cell 8 or 20 times. We generate three DARTS architecture datasets; each dataset contains 150 20-cell architectures randomly sampled from the search space but follows a different training set-up.  Specifically, for the dataset used in \textbf{Figure 2 (b) of main text}, we use an initial learning rate of 0.025, a cosine-annealing schedule and a batch size of 96 (i.e. the setting for CIFAR10 complete training in \cite{Liu2019_DARTS}. For the dataset used in \textbf{Figure 2 (c)  of main text}, we use an initial learning rate of 0.1, a step-decay schedule and a batch size of 128 (i.e. the setting for ImageNet complete training in \cite{Liu2019_DARTS}). Finally, for the dataset used in \textbf{Figure 2 (d) of main text}, we use an initial learning rate of 0.05, a cosine-annealing schedule and a batch size of 128, modified from the setting in Figure 2 (b). The other training setups are the same: all architectures are trained for 150 epochs using SGD optimiser with momentum of 0.9 and regularised using a Cut-Out of 16 and a DropPath with probability of 0.2 following \cite{Liu2019_DARTS}. For these datasets, we also record the training loss for each minibatch (TLmini) on top of the conventional training and validation loss/accuracies. The minibatch training loss is used to verify our claim that it is the sum of training losses, which has nice theoretical motivation, instead of training loss itself that gives good correlation with the generalisation performance of the architectures.
    
    \item \textbf{NASBench-301} \cite{siems2020bench}: To experiment on a larger number of architectures formed of DARTS cells, we further experiment on this dataset which contains 23000 8-cell architectures drawn from the DARTS search space and evaluated on CIFAR10. These architectures are much smaller than the 20-cell counterparts we generate and can be trained to convergence in fewer epochs. Each architecture in this dataset is trained for 100 epochs using a SGD optimiser with an initial learning rate of 0.025, a cosine annealing schedule and a batch size of 96. The training also adopts regularisation techniques such as an auxiliary tower with a weight of 0.4 and DropPath with probability of 0.2. For these architectures, we can assess their training accuracy/loss, validation accuracy after every training epoch from the dataset. Moreover, this dataset also provides a well trained surrogate model which can accurately predict the final test accuracy of any other possible architectures formed by 8 cells from the DARTS search space. We use this surrogate model to predict the ground-truth test accuracy of the subnetworks, when being trained from scratch independently, in the one-shot experiments in Section 4.4. The dataset is available at \url{https://github.com/automl/nasbench301} with Apache-2.0 License.
    
        \item \textbf{ResNet} \cite{Radosavovic2019}: It features two ResNet model families: ResNet and ResNeXt. The number of architecture samples as well as the architecture parameters and their corresponding range are shown in Table \ref{tab:resnet_search_space}.  Each architecture is trained on CIFAR10 for 100 epochs using SGD with an initial learning rate of 0.1, a cosine annealing schedule and a batch size of 128.  The dataset is available at \url{https://github.com/facebookresearch/nds} with  MIT License.
        \begin{table}[ht!b]
\caption{Search spaces for ResNet and ResNeXt. Each network is formed of three stages and for each of the stage $i$, there are $d_i$ the number of blocks,  $w_i$ number of channels per block. For ResNeXt, we also need to decide on the bottleneck width ratio $r_i$ and the number of groups $g_i$ per stage. The total number of possible architectures is $(dw)^3$ and $(dwrg)^3$ for ResNet and ResNeXt.}
\label{tab:resnet_search_space}
\centering
\begin{tabular}{@{}lllllll@{}}
\toprule
          & $d_i$      & $w_i$         & $r_i$     & $g_i$      & $N_{samples}$ & $N_{total}$ \\ \midrule
ResNet    & [1,24] & [16,256]  &           &            & 25000         & 1259712     \\
ResNeXt-A & [1,16] & [16,256]  & [1,4] & [1,4]  & 25000         & 11,390,625  \\
ResNeXt-B & [1,16] & [64,1024] & [1,4] & ]1,16] & 25000         & 52,734,375  \\ \bottomrule
\end{tabular}
\end{table}
        
\end{itemize}


\section{Training Losses vs Validation Losses} \label{app:tl_vs_vl}

\subsection{Compute TSE with training losses or validation losses} \label{app:tl_vs_vl_compare}

\begin{figure}[ht!b]
     \centering
    \begin{subfigure}{0.3\linewidth}
     \centering
    \includegraphics[trim=0cm 0.cm 0cm  0.1cm, clip, width=1.0\linewidth]{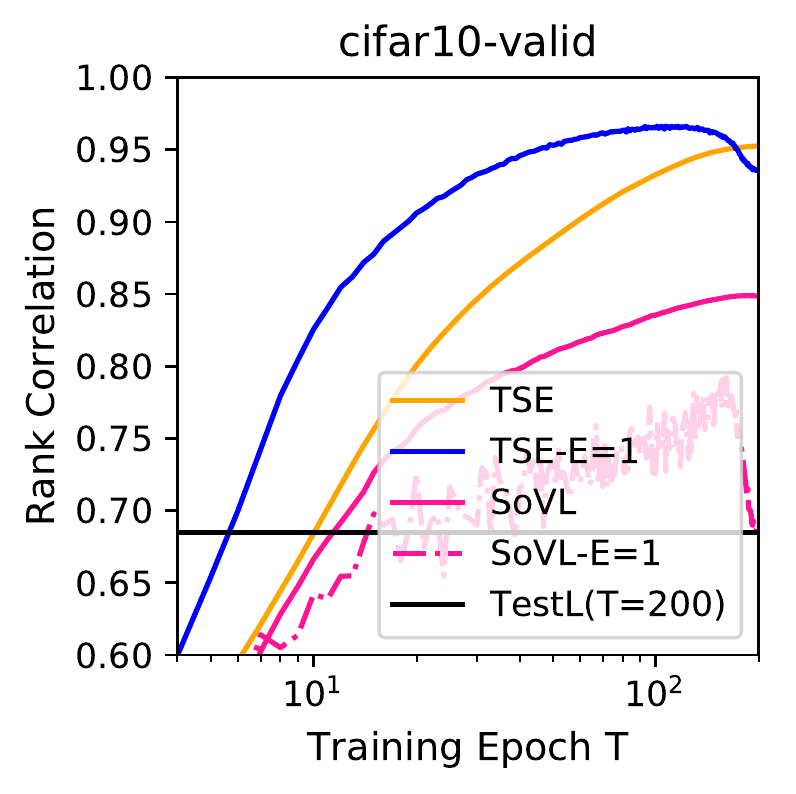}
    \caption{CIFAR10}
    \end{subfigure}
    \begin{subfigure}{0.3\linewidth}
     \centering
    \includegraphics[trim=0cm 0.cm 0cm  0.1cm, clip, width=1.0\linewidth]{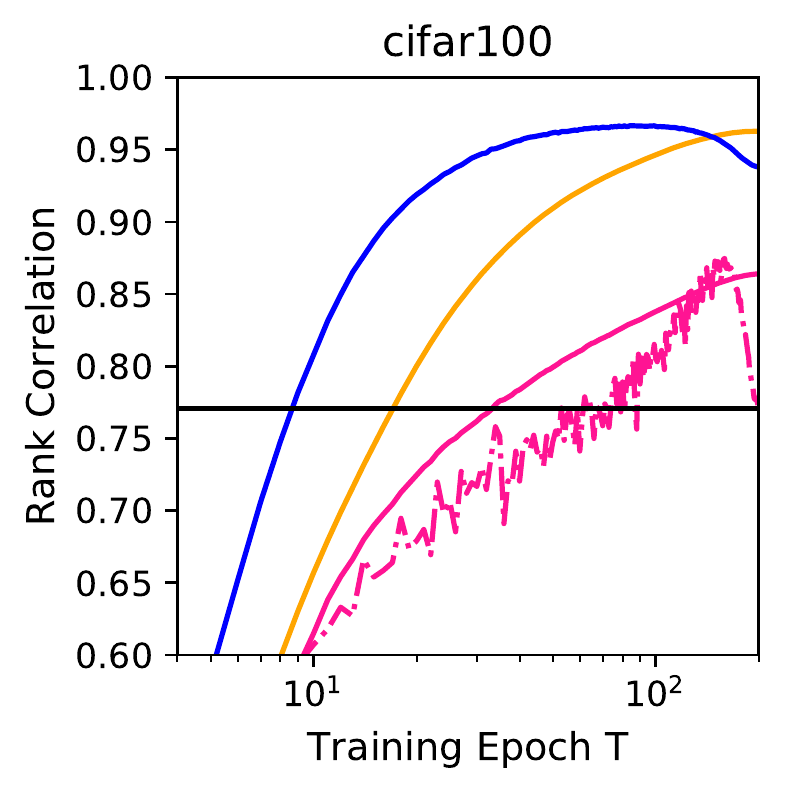}
    \caption{CIFAR100}
    \end{subfigure}
    \begin{subfigure}{0.3\linewidth}
     \centering
    \includegraphics[trim=0cm 0.cm 0cm  0.1cm, clip, width=1.0\linewidth]{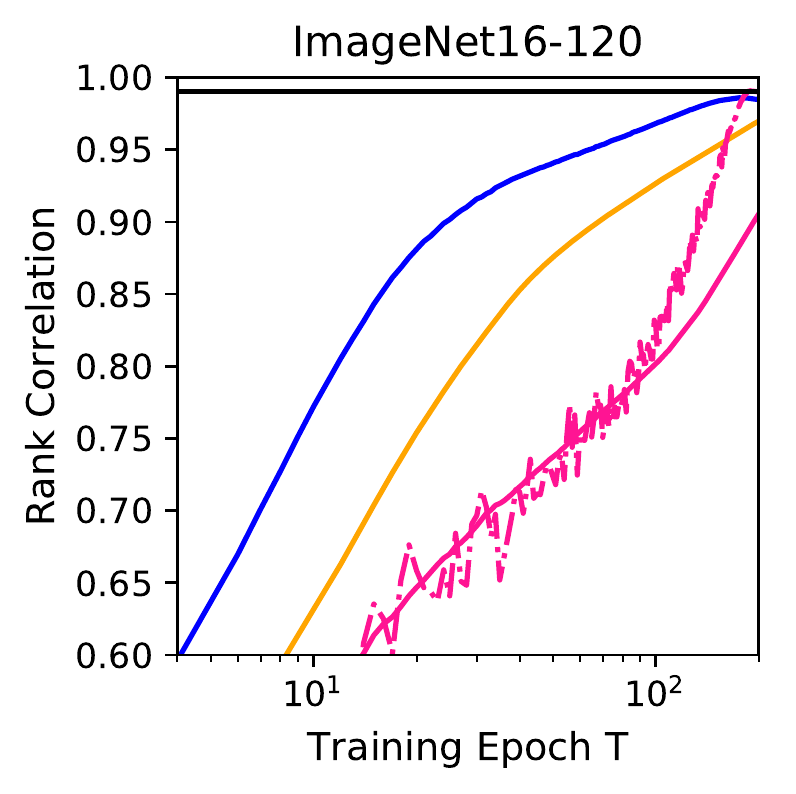}
    \caption{IMAGENET-16-120}
    \end{subfigure}
     \caption{Rank correlation (with final test \emph{accuracy}) performance of TSE (yellow) and TSE-E (blue), and those of validation losses (pink), SoVL (solid) and SoVL-E (dash dot), as well as that of final test loss (black) for architectures in NASBench-201 on three image datasets. Note the correlation performances of final test loss and SoVL-E near the end of training get surprisingly poor for CIFAR10/100. We explain this in Appendix \ref{app:tl_vs_vl_compare}.} \label{fig:sotl_vs_sovl}
\end{figure}

We perform a simple sanity check against the validation loss on NASBench-201 datasets. Specifically, we compare our proposed estimators, TSE and TSE-E, computed with training losses against two equivalent variants of validation loss-based estimators:  Sum of validation losses (SoVL) and that over the most recent epoch (SoVL-E with $E=1$). For each image dataset, we randomly sample 5000 different neural network architectures from the search space and compute the rank correlation between the true test accuracies (at $T=200$) of these architectures and their corresponding TSE/TSE-E as well as SoVL/SoVL-E up to epoch $T$. The results in Figure \ref{fig:sotl_vs_sovl} in the Appendix show that our proposed estimators TSE and TSE-E using training losses clearly outperform their validation counterparts. 

Another intriguing observation is that the rank correlation performance of SoVL-E drops significantly in the later phase of the training (after around 100 epochs for CIFAR10 and 150 epochs for CIFAR100) and the final test loss, TestL (T=200), also correlates poorly with final test \emph{accuracy}. This implies that the validation/test losses can become unreliable indicator for the validation/test accuracy on certain datasets; as training proceeds, the validation accuracy keeps improving but the validation losses could stagnate at a relatively high level or even start to rise \cite{mukhoti2020calibrating, soudry2018implicit}. This is because while the neural network can make more correct classifications on validation points (which depend on the maximum argument of the logits) over the training epochs, it also gets more and more confident on the correctly classified training data and thus the weight norm and maximum of the logits keeps increasing. 
This can make the network overconfident on the misclassified \emph{validation} data and cause the corresponding validation loss to rise, thus offsetting or even outweighing the gain due to improved prediction performance \cite{soudry2018implicit}.
Training loss will not suffer from this problem. 
While TSE-E struggles to distinguish architectures once their training losses have converged to approximately zero, this contributes to a much smaller drop in estimation performance of TSE-E compared to that of SoVL-E and only happens near a very late phase of training (after 150 epochs) which will hardly be reached if we want efficient NAS using as \emph{few} training epochs as possible. Therefore, the possibility of network overconfidence under misclassification is another reason for our use of training losses instead of the validation losses.   

\subsection{Example showing training loss is better correlated with validation accuracy than validation loss}

\begin{figure}[h!tb]
	\begin{subfigure}{0.3\linewidth}
     \centering
    \includegraphics[trim=0cm 0.cm 0cm  0.7cm, clip, width=1.0\linewidth]{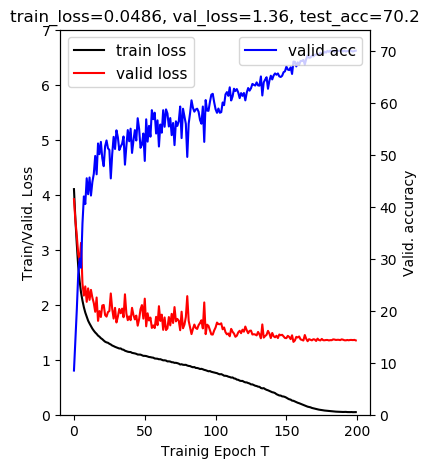}
    \caption{\centering Arch A: Train loss=0.05, Val. loss=1.36, Val. acc = 0.70}
    \end{subfigure}
    \begin{subfigure}{0.3\linewidth}
     \centering
    \includegraphics[trim=0cm 0.cm 0cm  0.7cm, clip, width=1.0\linewidth]{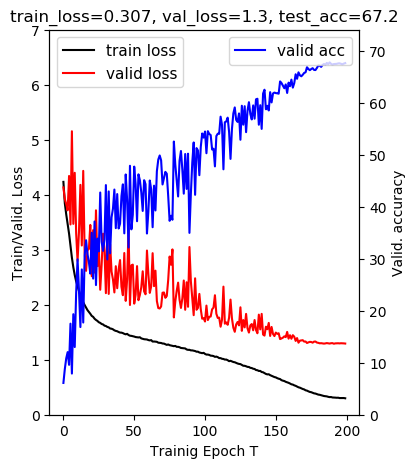}
    \caption{\centering Arch B: Train loss=0.31, Val. loss=1.30, Val. acc = 0.67}
    \end{subfigure}
    \begin{subfigure}{0.3\linewidth}
     \centering
    \includegraphics[trim=0cm 0.cm 0cm  0.7cm, clip, width=1.0\linewidth]{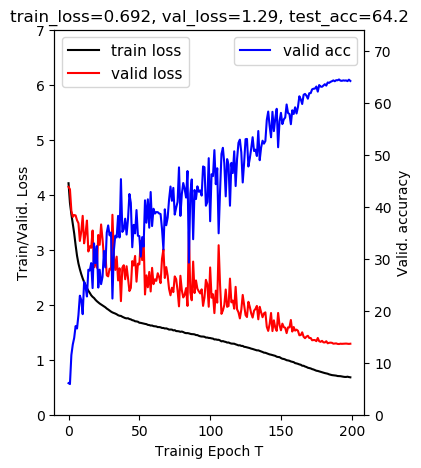}
    \caption{\centering Arch C: Train loss=0.69, Val. loss=1.29, Val acc = 0.64}
    \end{subfigure}
    \caption{Training losses, validation losses and validation accuracies of three example architectures on CIFAR100. The average of the training losses, validation losses and validation accuracies over the final 10 epochs is presented in the subcaption of each architecture.}\label{fig:3sample_archs_acc_loss}
\end{figure}

We sample three example architectures from the NASBench-201 dataset and plot their losses and validation accuracies on CIFAR100 over the training epochs $T$. The relative ranking for the validation accuracy is: Arch A (0.70) $>$ Arch B (0.67) $>$ Arch C (0.64), which corresponds perfectly (negatively) with the relatively ranking for the training loss: Arch A (0.05) $<$ Arch B (0.31) $<$ Arch C (0.69). Namely, the best performing architecture also has the lowest final training epoch loss. However, the ranking among their validation losses is poorly/wrongly correlated with that of validation accuracy; the worst-performing architecture has the lowest final validation losses but the best-performing architecture has the highest validation losses. Moreover, in all three examples, especially the better-performing ones, the validation loss stagnates at a relatively high value while the validation accuracy continues to rise. The training loss does not have this problem and it decreases while the validation accuracy increases. This confirms the observation we made in Appendix \ref{app:tl_vs_vl_compare} that the validation loss will become an unreliable predictor for the final validation accuracy as well as the generalisation performance of the architecture as the training proceeds due to overconfident misclassification. 

\begin{figure}[h!tb]
     \centering
    \begin{subfigure}{0.3\linewidth}
     \centering
    \includegraphics[trim=0cm 0.cm 0cm  0.1cm, clip, width=1.0\linewidth]{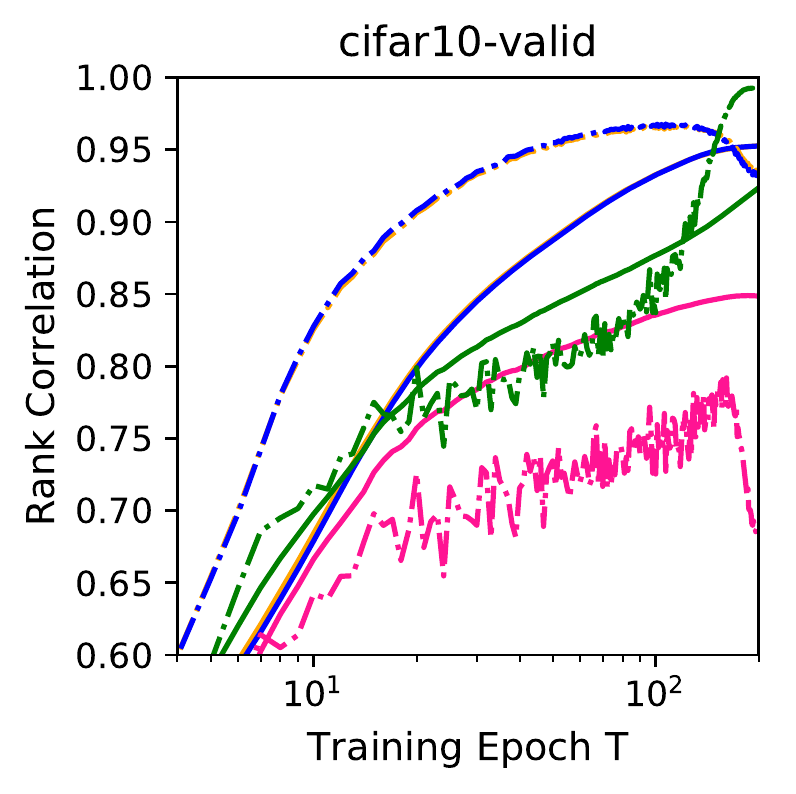}
    \caption{CIFAR10}
    \end{subfigure}
    \begin{subfigure}{0.3\linewidth}
     \centering
    \includegraphics[trim=0cm 0.cm 0cm  0.1cm, clip, width=1.0\linewidth]{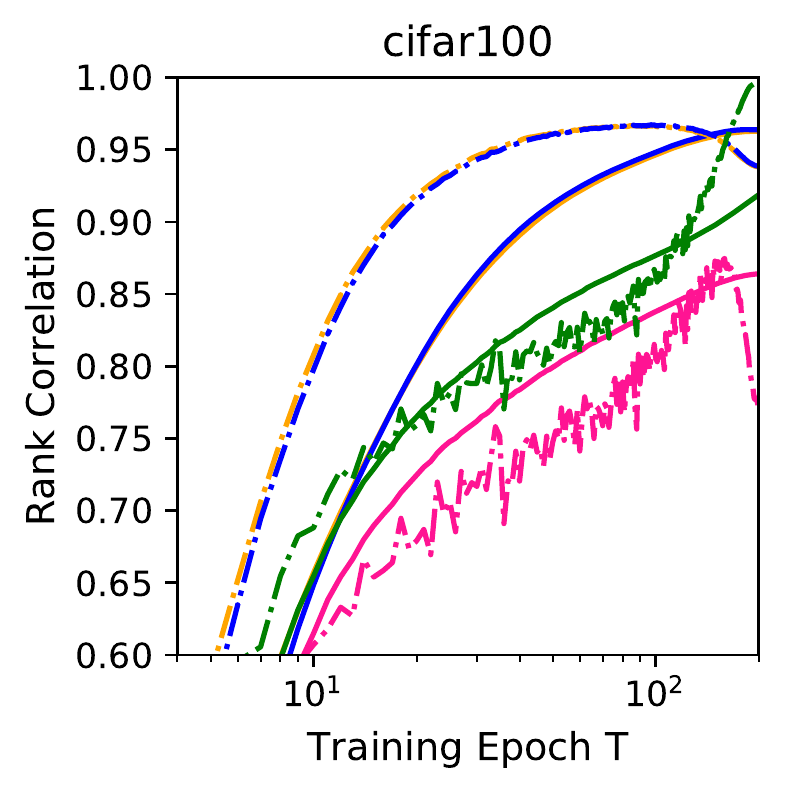}
    \caption{CIFAR100}
    \end{subfigure}
    \begin{subfigure}{0.3\linewidth}
     \centering
    \includegraphics[trim=0cm 0.cm 0cm  0.1cm, clip, width=1.0\linewidth]{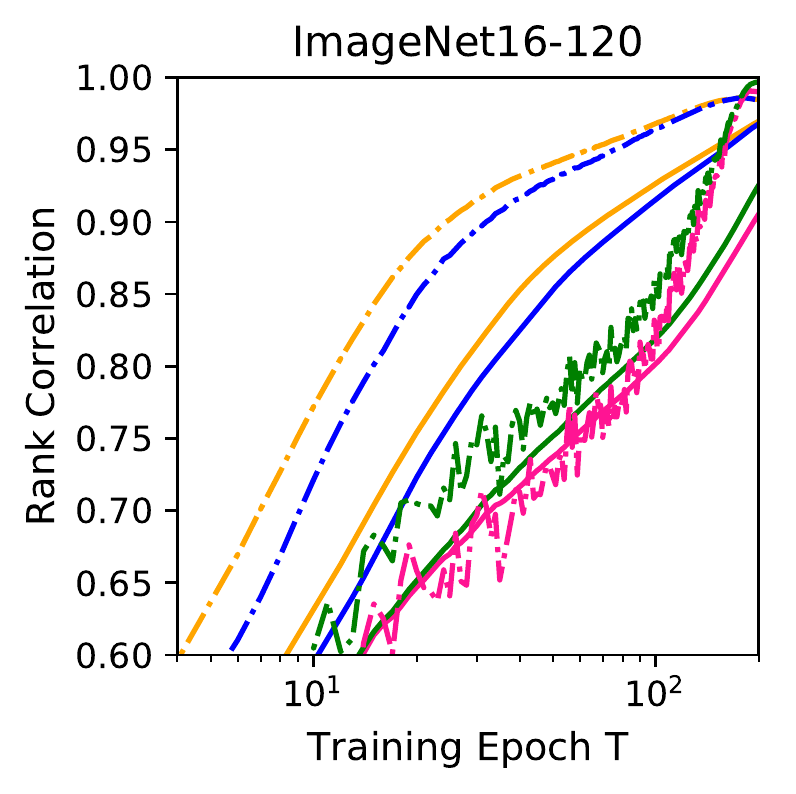}
    \caption{IMAGENET-16-120}
    \end{subfigure}
    \begin{subfigure}{1.0\linewidth}
    \includegraphics[trim=0cm 0.0cm 0cm  0cm, clip, width=1.0\linewidth]{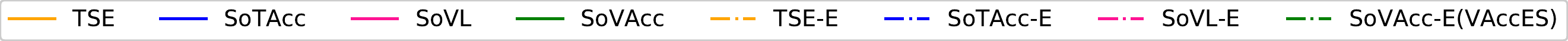}       \vspace{-0.5cm}
    \end{subfigure}
    \caption{Rank correlation performance of our TSE (yellow), the sum over training accuracy, SoTAcc (blue), the sum over validation losses, SoVL (pink), the sum of validation accuracy, SoVAcc (green) as well as their summing over the recent $E$ epoch counterparts (dash dot) for 5000 random architectures in NASBench-201 on three image datasets.} \label{fig:sotl_sovl_soac}
\end{figure}

\subsection{Comparison with sum over accuracy}

We compute another two variants of our estimator, SoTAcc and SoTAcc-E, by summing over training accuracies rather than training losses. Another two baselines to check against are the sum over validation accuracy, SoVAcc and SoVAcc-E. 
The results on CIFAR10 and CIFAR100 in Figure \ref{fig:sotl_vs_sovl} of the Appendix confirm the discussion in Appendix \ref{app:tl_vs_vl_compare}; as the training proceeds, the validation loss can become poorly correlated with the validation/test accuracy while the training loss is still perfectly correlated with the training accuracy. It is expected that SoVAcc-E should converge to a perfect rank correlation (=1) with the true test performance at the end of the training. However, the results in (a), (b) and (c) show that our proposed estimator \emph{TSE-E can consistently outperform SoVAcc-E} in the early and middle phase of the training (roughly $T \leq 150$ epochs). This reconfirms the usefulness of our estimator.

\subsection{Overfitting on CIFAR10 and CIFAR100}\label{subsec:overfit}

\begin{figure}[!htb]
     \centering
    \begin{subfigure}{0.45\linewidth}
     \centering
    \includegraphics[trim=0.4cm 0.3cm 8cm  0.0cm, clip, width=1.0\linewidth]{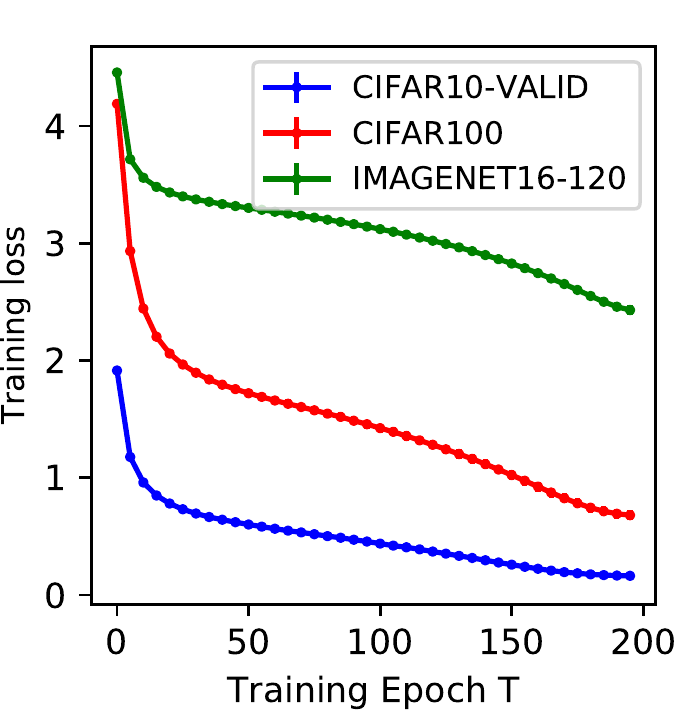}
    \caption{Training loss}\label{subfig:train_loss_3datasets}
    \end{subfigure}\hspace{5pt}
    \begin{subfigure}{0.48\linewidth}
     \centering
    \includegraphics[trim=7.5cm 0.3cm 0.4cm  0.0cm, clip, width=1.0\linewidth]{train_loss_stats_ImageNet16-120s888of_first5000_archs_test_mean_stdTrue.pdf}
    \caption{No. of arch with loss$<0.1$}  \label{subfig:n_overfit_archs}
    \end{subfigure}
    
\caption{Mean and $5$ standard error of training losses and validation losses on all architectures on different NASBench-201image datasets. (a) shows the training curves and (b) shows the number of architectures whose training losses go below 0.1 as the training proceeds. Many architectures reach very small training loss in the later phase of the training on CIFAR10 and CIFAR100, thus may overfitting on these two datasets. But all the architectures suffer high training losses on IMAGENET-16-120, which is a much more challenging classification task, and none of them overfits.}
\end{figure}

In Figure \ref{fig:sotl_vs_sovl} of Appendix \ref{app:tl_vs_vl_compare}, the rank correlation achieved by TSE-E on CIFAR10 and CIFAR100 drops slightly after around $T=150$ epochs but a similar trend is not observed for IMAGENET-16-120. We hypothesise that this is because many architectures converge to very small training losses on CIFAR10 and CIFAR100 in the later training phase, making it more difficult to distinguish between these good architectures based on their later-epoch training losses. However, this does not happen on IMAGENET-16-120 because it is a more challenging dataset. We test this by visualising the training loss curves of all architectures in Figure \ref{subfig:train_loss_3datasets} of the Appendix, where the solid line and error bar correspond to the mean and standard error, respectively. We also plot out the number of architectures with training losses below 0.1 \footnote{the threshold 0.1 is chosen following the threshold for optimisation-based measures in \cite{jiang2020fantastic}} in Figure \ref{subfig:n_overfit_archs} of the Appendix. It is evident that CIFAR10 and CIFAR100 both see an increasing number of overfitted architectures as the training proceeds whereas all architectures still have high training losses on IMAGENET-16-120 at end of the training $T=200$ and none have overfit. Thus, our hypothesis is confirmed. In addition, similar observation is also shared in \cite{jiang2020fantastic} where the authors find the number of optimisation iterations required to reach loss of 0.1 correlates well with generalisation but the number of iterations required to go from a loss of 0.1 to 0.01 does not. 

\section{TSE Estimator Hyperparameters} \label{app:add_nas_exps}
\begin{figure}[t]
     \centering
    \begin{subfigure}{0.24\linewidth}
     \centering
    \includegraphics[trim=0.2cm 0.2cm 0.3cm  0.7cm, clip, width=1.0\linewidth]{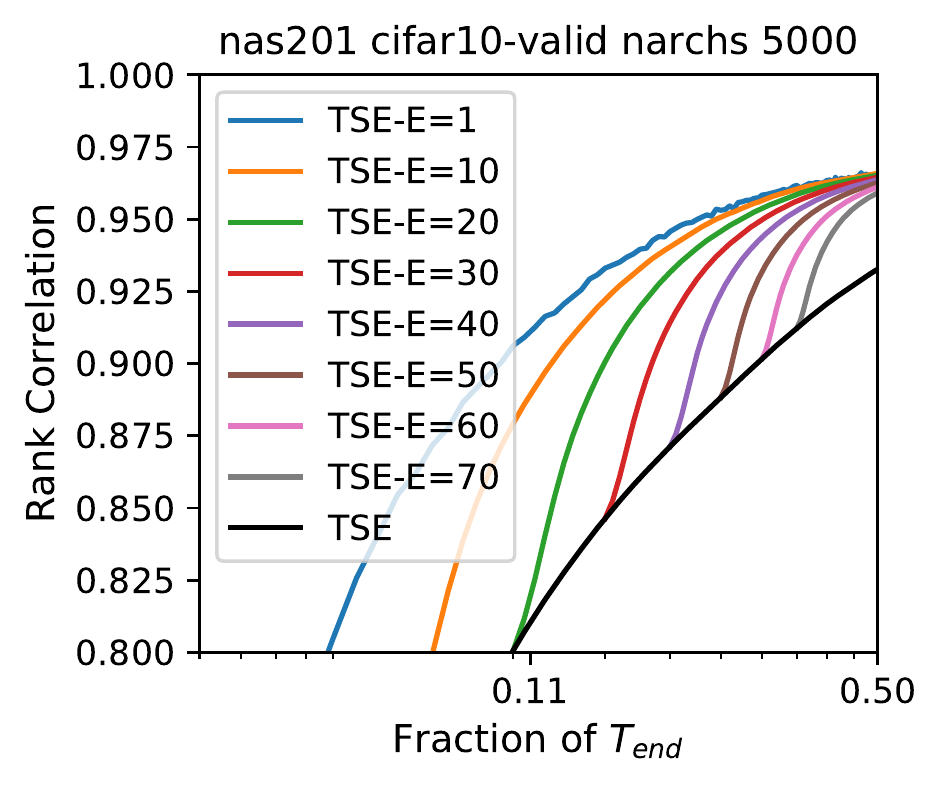}
    \caption{NB201-CIFAR10}
    \end{subfigure}
    \begin{subfigure}{0.24\linewidth}
     \centering
    \includegraphics[trim=0.2cm 0.2cm 0.3cm  0.7cm,, clip, width=1.0\linewidth]{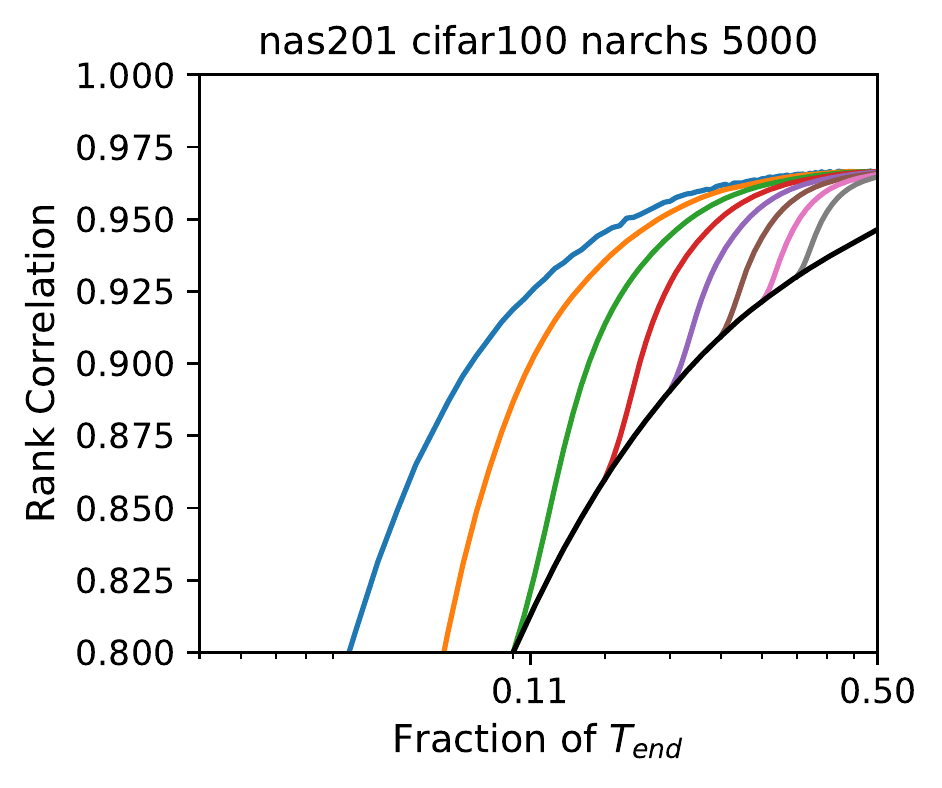}
    \caption{NB201-CIFAR100}
    \end{subfigure}
    \begin{subfigure}{0.24\linewidth}
     \centering
    \includegraphics[trim=0.2cm 0.2cm 0.3cm  0.7cm,, clip, width=1.0\linewidth]{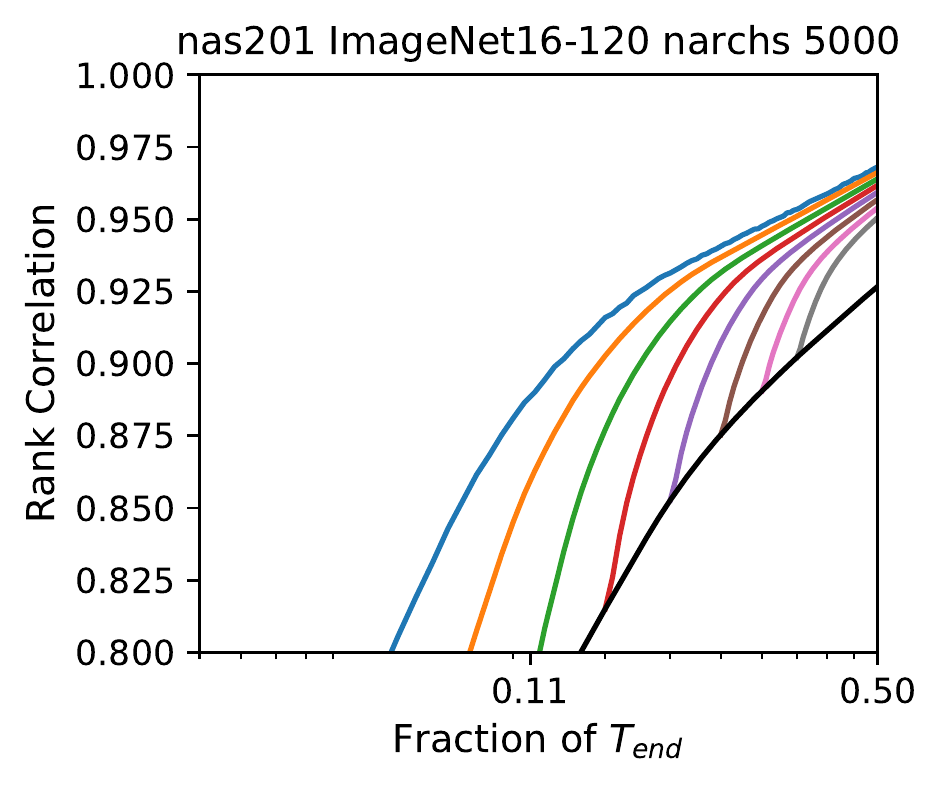}
    \caption{NB201-IMAGENET}
    \end{subfigure}
     \begin{subfigure}{0.24\linewidth}
     \centering
    \includegraphics[trim=0.2cm 0.2cm 0.3cm  0.7cm, clip, width=1.0\linewidth]{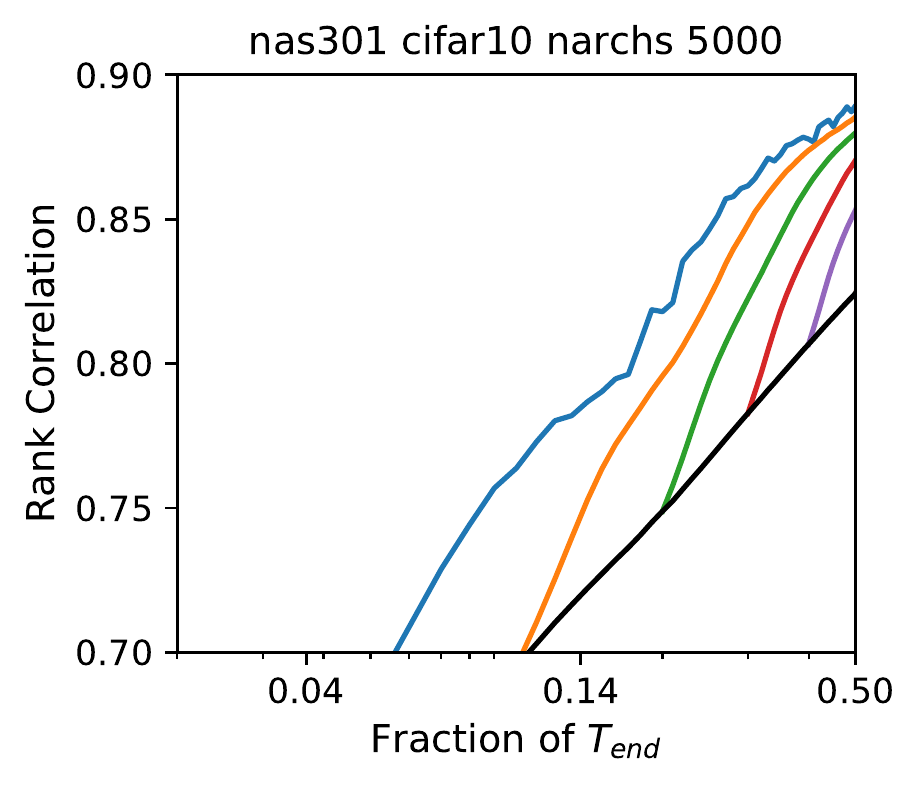}
    \caption{DARTS}
    \end{subfigure}
    \caption{Rank correlation performance of TSE-E computed over $E$ most recent epochs. Different $E$ values are investigated for architectures in NASBench-201 (NB201) on three image datasets and 5000 architectures from NASBench-301 (DARTS) on CIFAR10. In all cases, smaller $E$ consistently achieves better rank correlation performance with $E=1$ being the best choice.} \label{fig:window_size}
\end{figure}

\begin{figure}[t]
     \centering
    \begin{subfigure}{0.24\linewidth}
     \centering
    \includegraphics[trim=0.2cm 0.2cm 0.3cm  0.7cm, clip, width=1.0\linewidth]{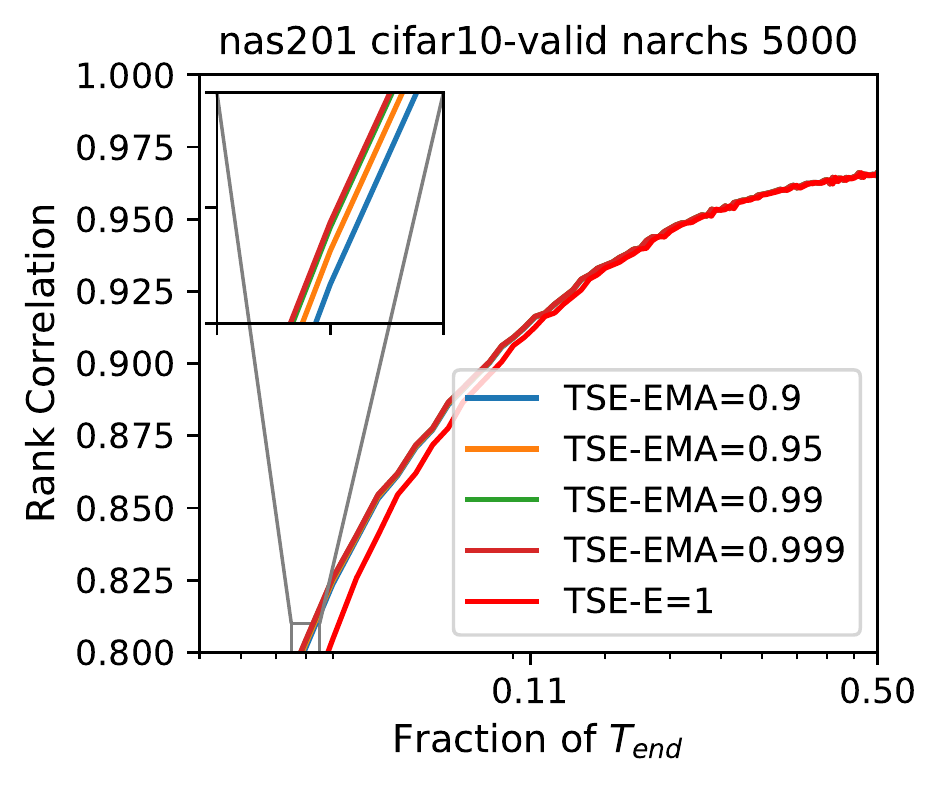}
    \caption{NB201-CIFAR10}
    \end{subfigure}
    \begin{subfigure}{0.24\linewidth}
     \centering
    \includegraphics[trim=0.2cm 0.2cm 0.3cm  0.7cm,, clip, width=1.0\linewidth]{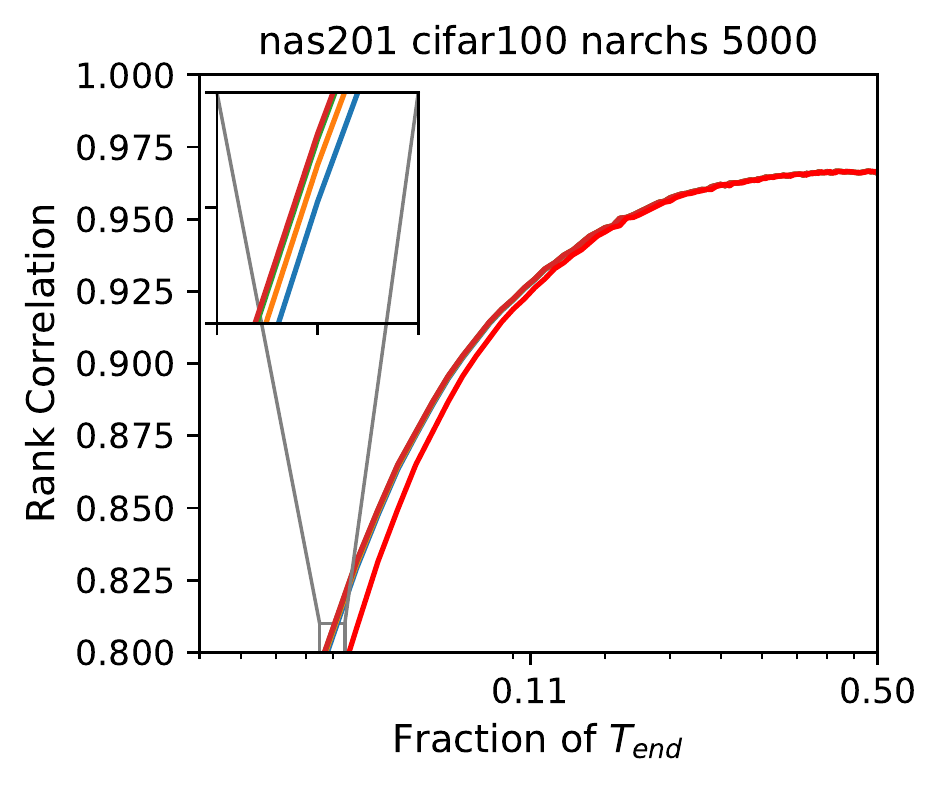}
    \caption{NB201-CIFAR100}
    \end{subfigure}
    \begin{subfigure}{0.24\linewidth}
     \centering
    \includegraphics[trim=0.2cm 0.2cm 0.3cm  0.7cm,, clip, width=1.0\linewidth]{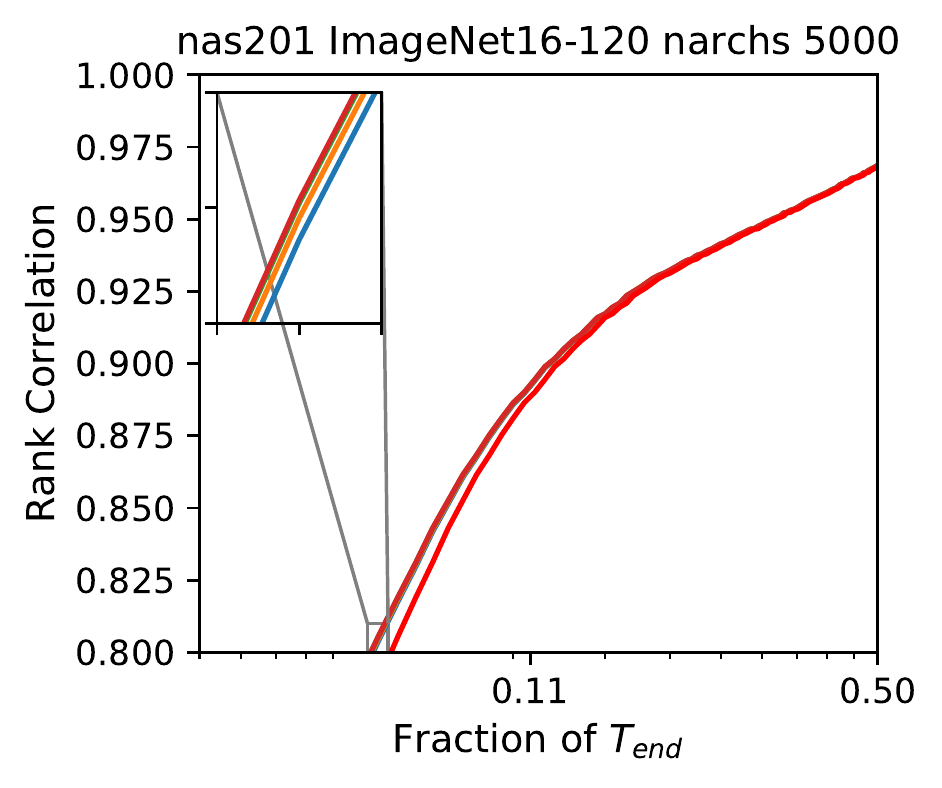}
    \caption{NB201-IMAGENET}
    \end{subfigure}
     \begin{subfigure}{0.24\linewidth}
     \centering
    \includegraphics[trim=0.2cm 0.2cm 0.3cm  0.7cm, clip, width=1.0\linewidth]{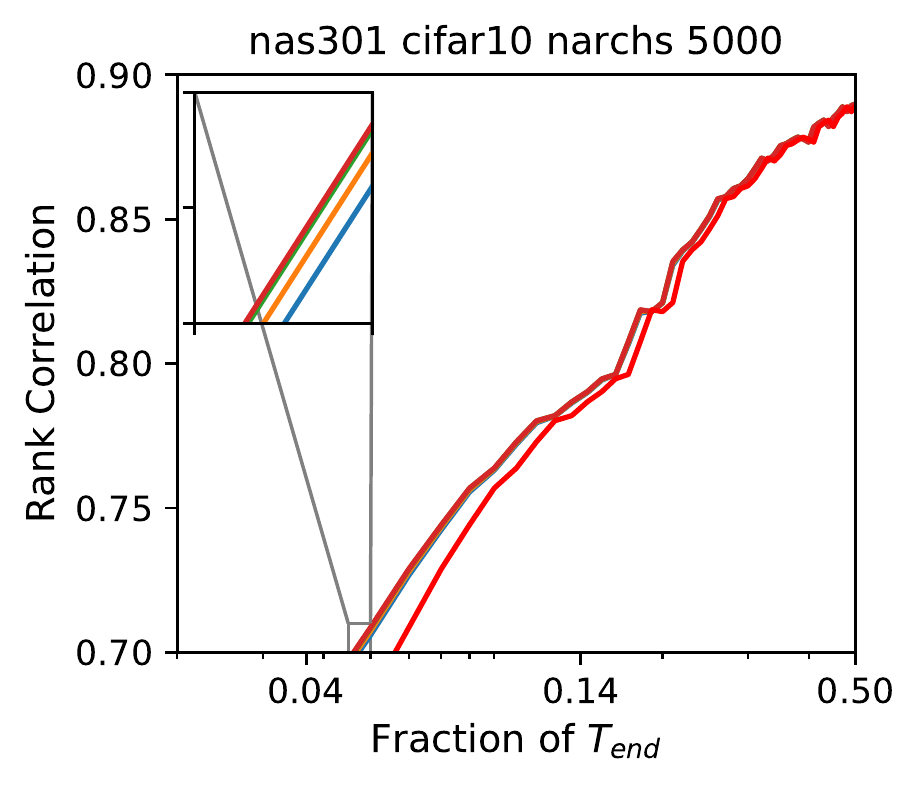}
    \caption{DARTS}
    \end{subfigure}
    \caption{Rank correlation performance of TSE-EMA with $\gamma$ on architectures in NASBench-201 (NB201) on three image datasets and 5000 architectures from NASBench-301 (DARTS) on CIFAR10. In all cases, TSE-EMA is very robust to different $\gamma$ values; the difference among TSE-EMA with different $\gamma$ is indistinguishable compared to that with TSE-E.} \label{fig:gamma}
\end{figure}

Our proposed TSE estimators require very few hyperparameters: the summation window size $E$ for TSE-E and the decay rate $\gamma$ for TSE-EMA, and we show empirically that our estimators are robust to these hyperparameters.
For $E$ of TSE-E, we test different summation window sizes on various search spaces and image datasets in  Figure \ref{fig:window_size} and find that $E=1$ consistently gives the best results across all cases. This, together with the almost monotonic improvement of our estimator's rank correlation score over the training budgets, supports our hypothesis discussed in Section \ref{sec:tse} that training information in the more recent epochs is more valuable for performance estimation. Note that TSE-E with $E=1$ corresponds to the sum of training losses over all the batches in one single epoch. 
As for $\gamma$, we show in Figure \ref{fig:gamma} that TSE-EMA is robust to a range of popular choices $\gamma \in [0.9, 0.95, 0.99, 0.999]$ across various datasets and search spaces. Specifically, the performance difference among these $\gamma$ values are almost indistinguishable compared to the difference between TSE-EMA and TSE-E. Thus, we set $E=1$ and $\gamma=0.999$ in all the following experiments and recommend them as the default choice for potential users who want to apply TSE-E and TSE-EMA on a new task without additional tuning.

\section{Architecture Generator Selection}\label{sub:model_selection}

\begin{figure}[t]
     \centering
     \begin{subfigure}{0.49\linewidth}
     \centering
    \includegraphics[trim=0.2cm 0.25cm 0.0cm  0.65cm, clip, width=1.0\linewidth]{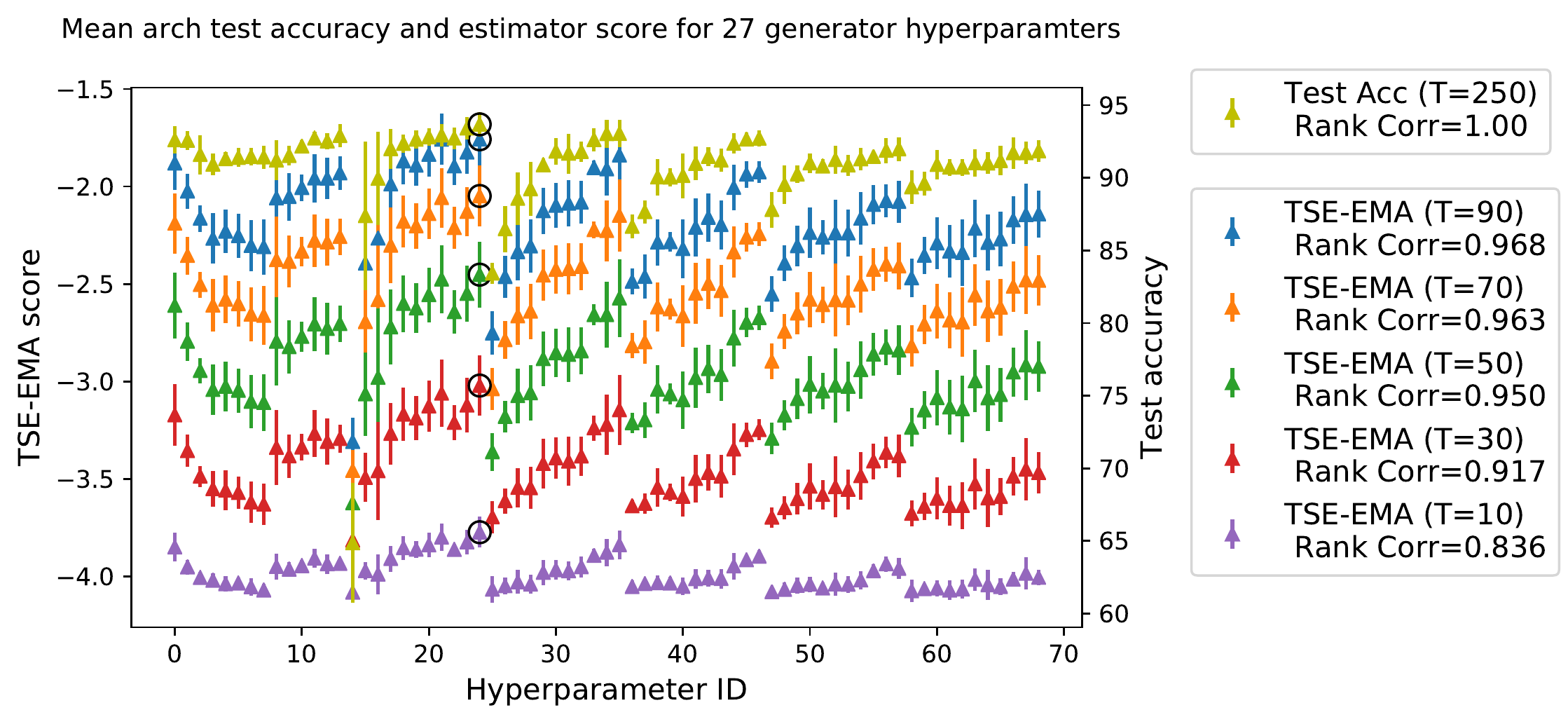}
    \caption{Training speed estimator: TSE-EMA}
    \end{subfigure}
     \begin{subfigure}{0.49\linewidth}
     \centering
    \includegraphics[trim=0.2cm 0.25cm 0.0cm  0.65cm, clip, width=1.0\linewidth]{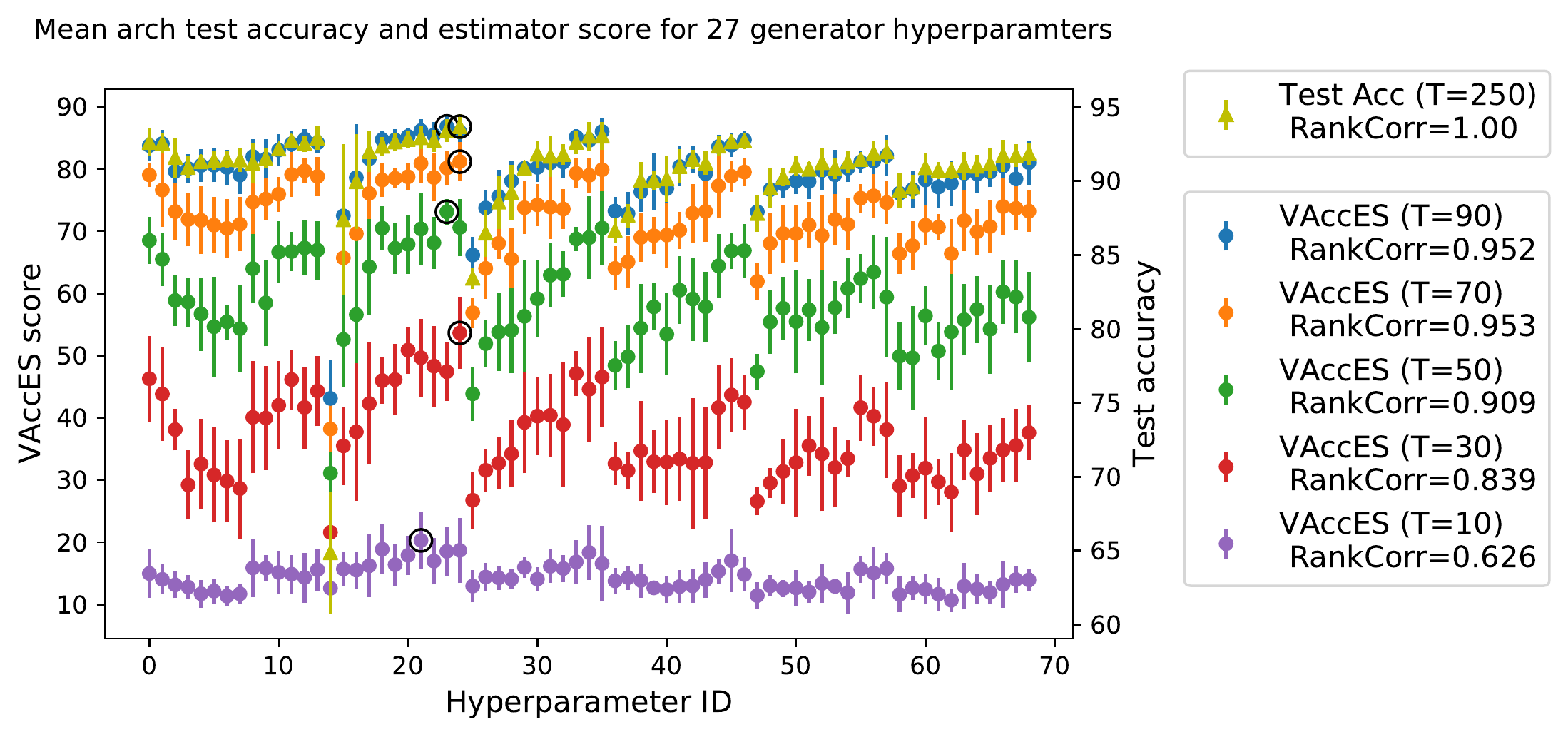}
    \caption{Early-stopped validation accuracy: VAccES}
    \end{subfigure}
    \vspace{-0.2cm}
    \caption{Model selection among $69$ random graph generator hyperparamters on RandWiredNN dataset using our TSE-EMA (a) and VAccES (b). We use each hyperparameter value to generate $8$ architectures and evaluate their true test accuracies after complete training. The mean and standard error of the test performance across $8$ architectures for each hyperparameter value are presented as Test Acc (yellow) and treated as ground truth (Right y-axis). We then compute our TSE-EMA estimator for all the architectures by training them for $T < T_{end}=250$ epochs. The mean and standard error of TSE-EMA scores for $T=10, \ldots, 90$ are presented in different colours (Left y-axis of (a)). The rank correlation between the mean Test Acc and that of TSE-EMA for various $T$ is shown in the corresponding legends in (a). The same experiment is conducted by using early-stopped validation accuracy (VAccES) for performance estimation (b). With only $10$ epochs of training, our TSE-EMA estimator can already capture the trend of the true test performance of different hyperparameters relatively well (Rank correlation$=0.851$) and can successfully identify 24-th hyperparamter setting as the optimal choice. The prediction of best hyperparameter by VAccES is less consistent and the rank correlation scores of VAccES at all epochs are lower than those of TSE-EMA.} \label{fig:model_selection}
\end{figure}

For the RandWiredNN dataset, we use $69$ different hyperparameter values for the random graph generator which generates the randomly wired neural architecture. Here we would like to investigate whether our estimator can be used in place of the true test accuracy to select among different hyperparameter values. For each graph generator hyperparameter value, we sample $8$ neural architectures with different wiring. The mean and standard error of the true test accuracies, TSE-EMA scores and early stopped validation accuracy (VAccES) over the $8$ samples are presented in Fig. \ref{fig:model_selection}. Our estimator can accurately predict the relative performance ranking among different hyperparameters (Rank correlation $\geq 0.84$) and accurately identify the optimal hyperparameter (circled in black) based on as few as $10$ epochs of training ($T=10$). The prediction by VAccES is less consistent and accurate and the rank correlation between VAccES and the final test accuracy is always lower than that of our TSE-EMA across different training budgets.

\section{Effective Training Budget for Our TSE Estimators}
\begin{figure}[t]
         \centering
	\begin{subfigure}{0.45\linewidth}
     \centering
    \includegraphics[trim=0.25cm 0.3cm 0.2cm  0.2cm, clip, width=0.8\linewidth]{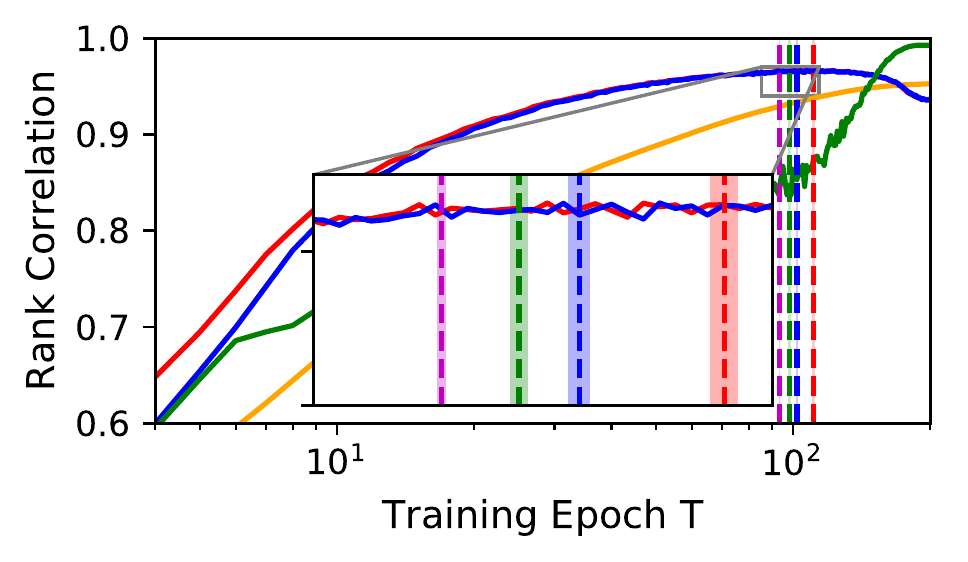}
        \caption{NB201-CIFAR10}
    \end{subfigure}
	\begin{subfigure}{0.45\linewidth}
     \centering
    \includegraphics[trim=0.25cm 0.3cm 0.2cm  0.2cm, clip, width=0.8\linewidth]{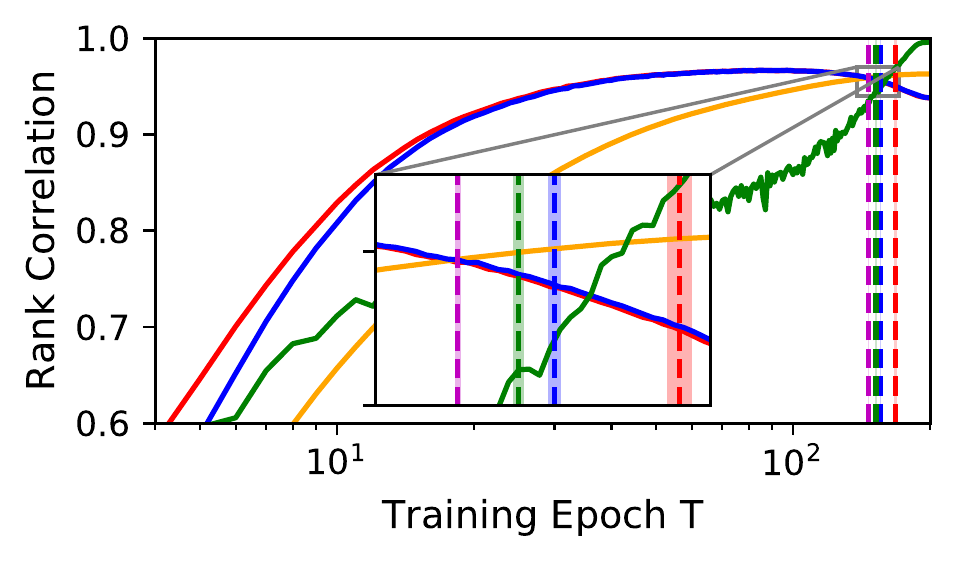}
    \caption{NB201-CIFAR100}
    \end{subfigure}
    	\begin{subfigure}{0.45\linewidth}
     \centering
    \includegraphics[trim=0.25cm 0.3cm 0.2cm  0.2cm, clip, width=0.8\linewidth]{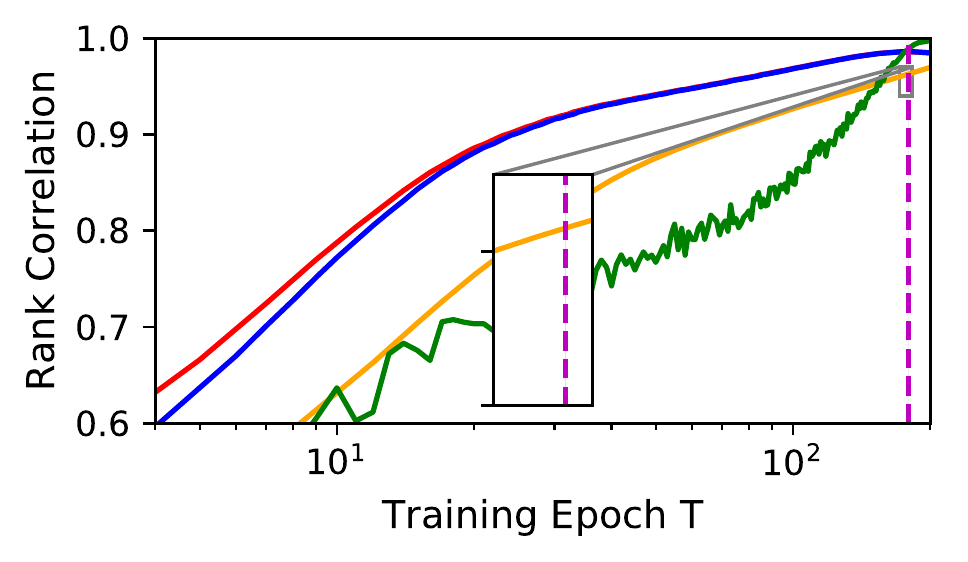}
    \caption{NB201-ImageNet}
    \end{subfigure}
         \begin{subfigure}{1.0\linewidth}
    \includegraphics[trim=0cm 0.0cm 0cm  0cm, clip, width=1.0\linewidth]{legend_termination.pdf}
    \end{subfigure}
    \caption{Rank correlation performance up to $T= T_{end}$. If the users want to apply our estimators for large training budget, they can estimate the effective range of our estimators based on the minimum epoch $T_{o}$ when overfitting happens among the $N_s$ observed architectures. They can then stop our estimators early at $0.9T_{o}$(marked by vertical lines) or switch back to validation accuracy beyond that. }\label{fig:termination_criterion_more}  
 \end{figure}

Our estimators can achieve superior rank correlation with the true generalisation performance for a relatively wide range of training budgets $T < T_{end}$.  However, our estimator is not meant to replace the validation accuracy at the end of training $T=T_{end}$ or when the user can afford large training budget to sufficient train the model. In those settings, validation accuracy remains as the gold standard for evaluating the true test performance of architectures. This is shown in Figure \ref{fig:termination_criterion_more} that as the training budget approaches $T_{end}$, our estimators will eventually be overtaken by validation accuracy. While the region requiring large training budget is less interesting for NAS where we want to maximise the cost-saving by using performance estimators, if the user does want to apply our estimators with a relatively large training budget, we propose a simple method here to estimate when our estimators would be less effective than validation accuracy.

\begin{algorithm}[ht!b]
	    \caption{Find Effective Training Budget for TSE Estimators}\label{alg:check_effect_T}
	\begin{algorithmic}[1]
		\STATE {\bfseries Input:}  A subset of $N_s$ fully trained architectures whose training losses are $\{ \{ \ell_{i,t} \}_{t=1}^{T_{end}} \}_{i=1}^{N_s}$, Overfitting criterion $is\_overfit()$
		\STATE {\bfseries Output:} The effective training budget $T_{effective}$ to use TSE-EMA or TSE-E
	\STATE $\mathcal{S}_{T_o} =\emptyset$
        \FOR{$i=1, \dots, N_s$}
        \FOR{$t=1, \dots, T_{end}$}
	\IF{$is\_overfit(\ell_{i,t}) ==$ True}
		\STATE $T_{i,o} = T$
		\STATE break
	\ELSE
		\STATE $T_{i,o} = T_{end}$
	\ENDIF
    	\ENDFOR
	\STATE $\mathcal{S}_{T_o} = \mathcal{S}_{T_o} \cup T_{i,o}$
	\ENDFOR
 	\STATE $T_o = \min \mathcal{S}_{T_o} $
	\STATE $T_{effective} = 0.9 T_o$
	\end{algorithmic}
\end{algorithm}

As discussed in Appendix \ref{subsec:overfit}, we notice that that our estimators, TSE-EMA and TSE-E, become less effective when the architectures compared start to overfit because both of them rely heavily on the lastest-epoch training losses to measure training speed, which is difficult to estimate when the training losses become too small. 
Thus, we can adopt the heuristic described in Algorithm \ref{alg:check_effect_T} to decide when to stop the computation of TSE-E and TSE-EMA early and revert to a previous checkpoint i.e. $T_{effective}$ \footnote{We use $T=0.9 T_{o}$ rather than $T_o$ to be conservative.}. Similar to Appendix \ref{subsec:overfit}, we use a simple criterion: training loss decreasing below $0.1$ (i.e. $\ell_{i,t=T_o} < 0.1$), to identify when an architecture start to overfit. We experiment with $N_s=10, 50, 100, 500$ architectures and the mean and standard error results over 100 random seeds are shown in Figure \ref{fig:termination_criterion_more}. It's evident that we can find a fairly reliable threshold with a sample size as small as $N_s=10$. 

\section{Additional NAS experiments} \label{app:add_nas_exps}

\subsection{Query-based NAS}
In this work, we incorporate our estimator, TSE-EMA, at $T=10$ into three NAS search strategies: Regularised Evolution \cite{real2019regularized}, Bayesian optimisation \cite{bergstra2011algorithms} and Random Search \cite{bergstra2012random} and performance architecture search on NASBench-201 datasets. We modify the implementation available at \url{https://github.com/automl/nas_benchmarks} for these three methods.

Random Search \cite{bergstra2012random} is a very simple yet competitive NAS search strategy \cite{Dong2020nasbench201}. We also combined our estimator, TSE-EMA, at training epoch $T=10$ with Random Search to perform NAS. We compare it against the baselines using the final validation accuracy at $T = 200$, denoted as Val Acc (T=200), and the early-stop validation accuracy at $T = 10$, denoted as Val Acc (T=10). Other experimental set-ups follow Section 4.3. The results over running hours on all three image tasks are shown in Figure \ref{fig:rs_nas} of the Appendix. Note the x-axis is in log scale. The use of our estimator clearly leads to faster convergence as compared to the use of final validation i.e. Val Acc (T=200). Moreover, our estimator also slightly outperforms the early-stop validation accuracy, Val Acc (T=10) on the three image tasks. The performance gain of using our estimator or the early-stopped validation accuracy is relatively less significant in the case of Random Search compared to the cases of Regularised Evolution and TPE. For example, given a budget of 50 hours on CIFAR100, Regularised Evolution and TPE when combined with our estimator can find an architecture with a test error around or below 0.26 but Random Search only finds architecture with test error of around 0.268. This is due to the fact that Random Search is purely explorative while Regularised Evolution and TPE both trade off exploration and exploitation during their search; our estimator by efficiently estimating the final generalisation performance of the architectures will enable better exploitation. Therefore, we recommend the users to deploy our proposed estimator onto search strategies which involve some degree of exploitation to maximise the potential gain.  

\begin{figure}[t]
	\begin{subfigure}{0.33\linewidth}
     \centering
    \includegraphics[trim=0cm 0.cm 0cm  0.3cm, clip, width=1.0\linewidth]{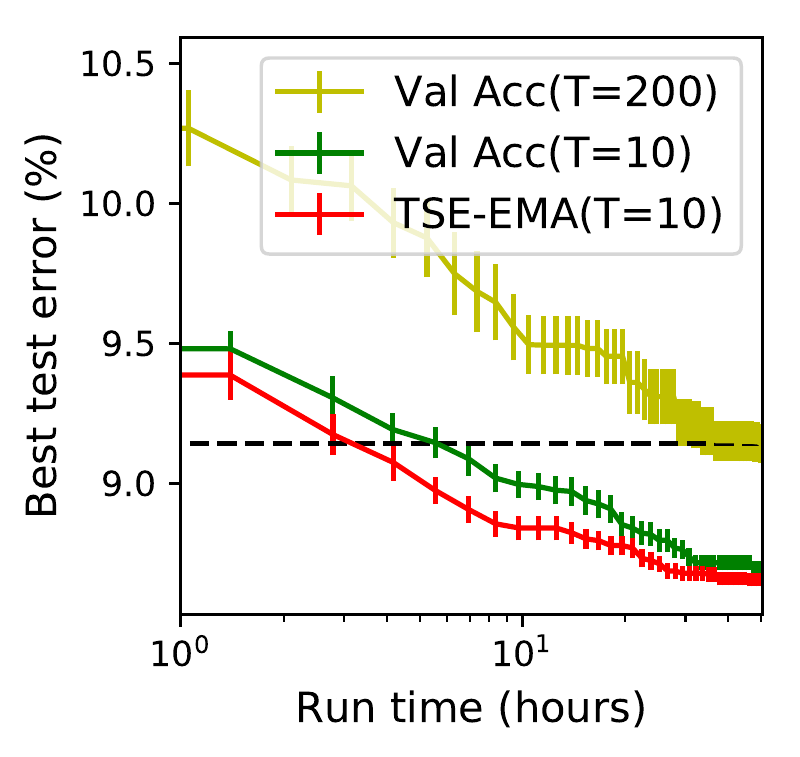}
    \caption{RS-CIFAR10}
    \end{subfigure}
    \begin{subfigure}{0.33\linewidth}
     \centering
    \includegraphics[trim=0cm 0.cm 0cm  0.3cm, clip, width=1.0\linewidth]{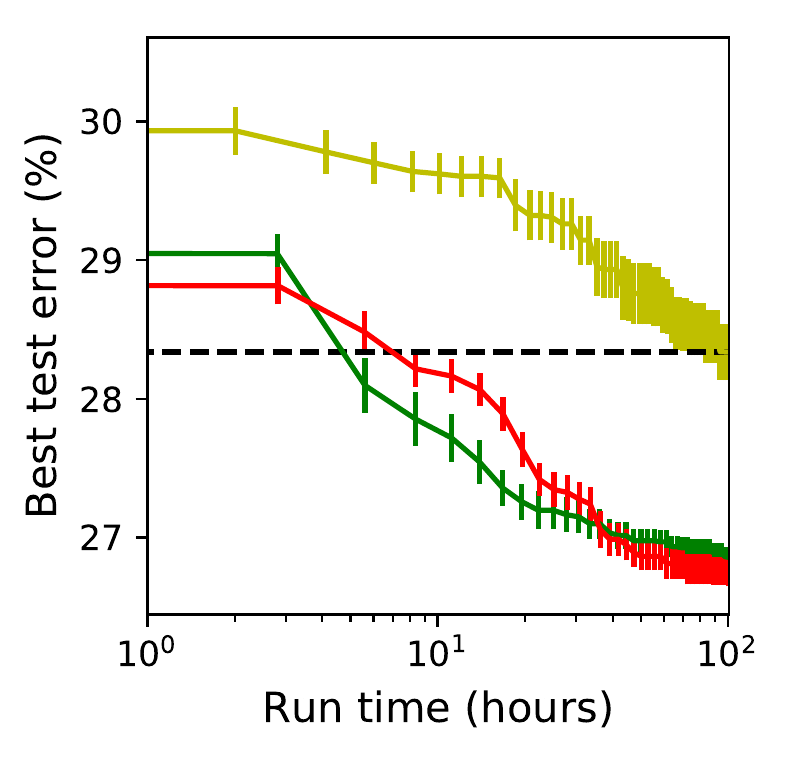}
    \caption{RS-CIFAR100}
    \end{subfigure}
    \begin{subfigure}{0.33\linewidth}
     \centering
    \includegraphics[trim=0cm 0.cm 0cm  0.3cm, clip, width=1.0\linewidth]{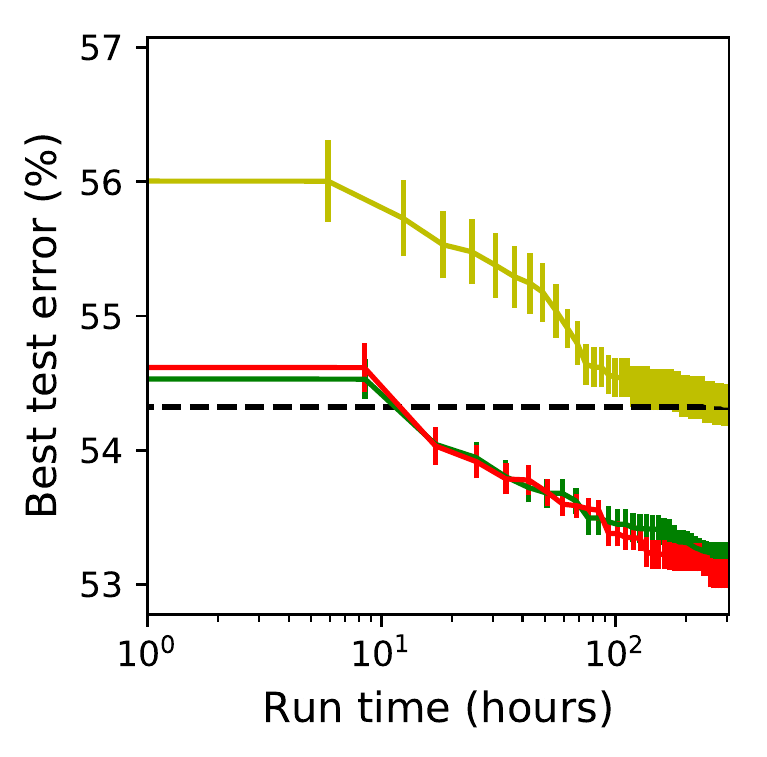}
    \caption{RS-IMAGENET-16-120}
    \end{subfigure}
    \caption{NAS performance of Random Search (RS) in combined with final validation accuracy (Final Val Acc), early-stop validation accuracy (ES Val Acc) and our estimator TSE-EMA on NASBench-201. TSE-EMA enjoys competitive convergence as ES Val Acc and both are faster than using Final Val Acc.}\label{fig:rs_nas}
\end{figure}

\subsection{One-shot NAS}

We follow the RandNAS\cite{li2020random} procedure for the supernetwork training but modify the search phase: for each randomly sampled subnetwork, we train it for $B$ additional mini-batches after inheriting weights from the trained supernetwork to compute our TSE estimator. Note although this introduces some costs, our estimator saves all the costs from evaluation on validation set as it doesn't require validation data.
To ensure fair comparison, we also recompute the validation accuracy, SoVL, Tlmini of each subnetwork after the additional training. We also experiment with more advanced supernetwork training techniques such as FairNAS \cite{chu2019fairnas} and MultiPaths \cite{yu2019universally} and show that our estimators can be applied on top of them to further improve the rank correlation performance. Please refer to Section 4.4 for more experimental details and to the start of Section 4 for description of different estimators. The complete table of results with more estimators are presented in Table \ref{tab:one_shot_nas_ext} in Appendix. 

\begin{table}[ht!b]
\caption{Results of performance estimators in one-shot NAS setting over 3 supernetwork training initialisations. For each supernetwork, we randomly sample 500 random subnetworks for DARTS and 200 for NB201, and compute their TSE, Val Acc, SoVL, Tlmini after inheriting the supernetwork weights and training for $B$ additional minibatches. Rank correlation measures the estimators' correlation with the rankings of the true test accuracies of subnetworks when \emph{trained from scratch independently}, and we compute the average test accuracy of the top 10 architectures identified by different estimators from all the randomly sampled subnetworks.}
\label{tab:one_shot_nas_ext}
    	\vspace{2mm} \resizebox{1.0\linewidth}{!}{
\begin{tabular}{@{}cccccccccc@{}}\toprule
\multirow{3}{*}{B}   & \multirow{3}{*}{Estimator} & \multicolumn{4}{c}{Rank Correlation}                                                     & \multicolumn{4}{c}{Average Accuracy of Top 10 Architectures}                                 \\\cmidrule(lr){3-6}\cmidrule(l){7-10}
                     &                            & \multicolumn{3}{c}{NB201-CIFAR10}                                         & DARTS               & \multicolumn{3}{c}{NB201-CIFAR10}                                      & DARTS                \\\cmidrule(lr){3-5}\cmidrule(l){6-6} \cmidrule(l){7-9}\cmidrule(l){10-10}
                     &                            & RandNAS              & FairNAS              & MultiPaths           & RandNAS             & RandNAS             & FairNAS             & MultiPaths          & RandNAS              \\\midrule
\multirow{2}{*}{100} & TSE                        & \textbf{0.70 (0.02)} & \textbf{0.84 (0.01)} & \textbf{0.83 (0.01)}         & \textbf{0.30(0.04)} & \textbf{92.67 (0.12} & \textbf{92.70 (0.10)} & 92.63 (0.12)& \textbf{93.64(0.04)} \\
                     & Val Acc                    & 0.44 (0.15)          & 0.56 (0.17)          & 0.67 (0.05)          & 0.11(0.04)          & 91.47 (0.31)          & 91.73 (0.21)        & 91.77 (0.78)          & 93.20(0.04)          \\
                     & SoVL                       & 0.54 (0.13)          & \textbf{0.84 (0.06)}          & \textbf{0.83 (0.01)} & 0.10(0.04)          & 92.57 (0.15)          & 92.67 (0.06)        & \textbf{92.73 (0.06)} & 93.21(0.04)          \\
                     & Tlmini                     & 0.62 (0.03)          & 0.72 (0.09)          & 0.74 (0.02)          & 0.05(0.03)          & 91.80 (0.40)          & 92.33 (0.40)        & 92.43 (0.06)          & 93.38(0.03)          \\
                     \midrule
\multirow{2}{*}{200} & TSE                        & \textbf{0.70 (0.03)} &      \textbf{0.850 (0.01)}       & \textbf{0.83 (0.01)} & \textbf{0.32(0.04)} & \textbf{92.70 (0.00)} &    \textbf{92.77 (0.06)}   & \textbf{92.73 (0.06)} & \textbf{93.55(0.04)} \\
                     & Val Acc                    & 0.41 (0.10)          &          0.56 (0.17)        & 0.53 (0.11)          & 0.09(0.02)          & 91.53 (0.55)          &   92.40 (0.10)      & 92.23 (0.23)          & 93.34(0.02)          \\
                     & SoVL                       & 0.52 (0.17)          &     0.84 (0.06)                 & 0.80 (0.02)          & 0.08(0.02)          & 90.70 (1.35)          &     92.53 (0.15)                & 92.50 (0.10)          & 93.36(0.02)          \\
                     & Tlmini                     & 0.46 (0.14)          &     0.72 (0.10)              & 0.69 (0.06)         & 0.02(0.01)          & 92.00 (0.35)          &      92.53 (0.25)               & 92.40 (0.27)          & 93.15(0.01)          \\
                     \midrule
\multirow{2}{*}{300} & TSE                        & \textbf{0.71 (0.03)} &   \textbf{0.85 (0.00)}              & \textbf{0.82 (0.01)}          & \textbf{0.34(0.04)} & \textbf{92.70 (0.00)} &      \textbf{92.77 (0.06)}    & \textbf{92.70 (0.00)} & \textbf{93.65(0.04)} \\
                     & Val Acc                    & 0.44 (0.04)          &        0.62 (0.08)              & 0.59 (0.71)          & 0.06(0.02)          & 91.20 (0.35)          &          92.10 (0.50)          & 91.43 (0.72)          & 93.31(0.02)          \\
                     & SoVL                       & 0.45 (0.21)          &      0.81 (0.05)                & 0.81 (0.03) & 0.05(0.02)          & 91.00 (1.60)          &      92.53 (0.15)                & 92.53 (0.06)          & 93.26(0.02)          \\
                     & Tlmini                     & 0.47 (0.12)          &    0.74 (0.02)                  & 0.70 (0.04)         & 0.09(0.01)          & 91.60 (0.44)         &      92.43 (0.21)               & 92.27 (0.12)          & 92.95(0.01) \\
                     \bottomrule         
\end{tabular}}\vspace{-0.2cm}
\end{table}

\subsection{Differentiable NAS} \label{app:tse_for_diff_nas}

We modify two differentiable approach, DARTS \cite{Liu2019_DARTS} and DrNAS \cite{chen2021drnas}, by directly using the derivative of our TSE estimator instead of that of the original validation loss to update the architecture (distribution) parameters. 

In DARTS \cite{Liu2019_DARTS}, each intermediate node $\phi^{(j)}$is computed based on all of its predecessors: 
\begin{equation}
	\phi^{(j)} = \sum_{i < j} \bar{o}^{(i,j)}\left(\phi^{(i)}\right)
\end{equation}
where $\bar{o}^{(i,j)}(\phi)$ is a mix of all possible operations  $o(\phi)$:
\begin{equation}
	\bar{o}^{(i,j)} \left(\phi \right) = \sum_{o \in \mathcal{O}} \frac{\exp (\alpha_o^{(i,j)})}{\sum_{o' \in \mathcal{O}} \exp (\alpha_{o'}^{(i,j)})} o(\phi)
\end{equation}
The architecture parameters to be searched in DARTS is thus a set of continuous vectors $\alpha = \{ \alpha^{(i,j)} \}$. Assume we run the search for $T$ epochs and each epoch comprises $B$ mini-batches, the algorithms of DARTS and DARTS-TSE is summarised in Algorithm \ref{alg:darts} and \ref{alg:darts_tse}. Note in DARTS, the architecture parameters are updated with the derivative of validation loss $\triangledown_{\alpha} \ell_{val}(w, \alpha)$ at each mini-batch, leading to a total of $BT$ updates. In DARTS-TSE, we compute the TSE estimator and its derivative using $K=100$ mini-batches, leading to a less frequent update of $\alpha$. To compensate that, we set $\triangledown_{\alpha} \ell_{TSE} \left(w, \alpha \right) = \sum_{k=1}^K \triangledown_{w} \ell^{(k)}_{train} \left(w, \alpha \right)$ instead of $\triangledown_{\alpha} \ell_{TSE} \left(w, \alpha \right) = \frac{1}{K} \sum_{k=1}^K \triangledown_{w} \ell^{(k)}_{train} \left(w, \alpha \right)$ for updating $\alpha$. 

%

DrNAS \cite{chen2021drnas} is very similar to DARTS but instead of updating the architecture parameters $\alpha$ directly, DrNAS assumes $\alpha$ is drawn from a Dirichlet distribution $q(\alpha \vert \beta)$ and optimise the distribution parameters $\beta$. $\beta$ is updated at each mini-batch by descending: 
\begin{equation} 
	\mathbb{E}_{q(\alpha \vert \beta)} \left[ \triangledown_{\beta} \ell_{val} \left(w, \alpha \right) \right]
\end{equation}
In DrNAS-TSE, we instead use the derivative of TSE to update $\beta$:
\begin{equation} 
\sum_{k=1}^B \mathbb{E}_{q(\alpha \vert \beta)} \left[ \triangledown_{\beta} \ell^k_{train} \left(w, \alpha \right) \right]
\end{equation}

For both DARTS and DrNAS, we set $K=100$ for computing our TSE and follow the default setting in \cite{Dong2020nasbench201,Liu2019_DARTS, chen2021drnas} for all the other hyperparameters including $B$ and $T$.

\begin{algorithm}[t]
	\caption{DARTS}\label{alg:darts}
	\begin{algorithmic}[1]
	\STATE Create a mixed operation $\bar{o}^{i,j}$ parametrised by $\alpha^{i,j}$ for each edge $(i,j)$
        \FOR{$t=1, \dots, BT$}
		\STATE Update architecture parameter $\alpha$ by descending $\triangledown_{\alpha} \ell_{val} \left(w, \alpha \right)$
		\STATE Update weights $w$ by descending $\triangledown_{w} \ell_{train} \left(w, \alpha \right)$
	\ENDFOR
 	\STATE Derive the final architecture based on the learned $\alpha$
	\end{algorithmic}
\end{algorithm}
\begin{algorithm}[t]
	\caption{DARTS-TSE}\label{alg:darts_tse}
	\begin{algorithmic}[1]
	\STATE Create a mixed operation $\bar{o}^{i,j}$ parametrised by $\alpha^{i,j}$ for each edge $(i,j)$
        \FOR{$t=1, \dots, \floor{BT/K}$}
		\STATE Update architecture parameter $\alpha$ by descending $\triangledown_{\alpha} \ell_{TSE} \left(w, \alpha \right)$
		\STATE $\triangledown_{\alpha} \ell_{TSE} \left(w, \alpha \right) = 0$
		\FOR{$k=1, \dots, K$}
			\STATE Update weights $w$ by descending $\triangledown_{w} \ell_{train} \left(w, \alpha \right)$
			\STATE $\triangledown_{\alpha} \ell_{TSE} \left(w, \alpha \right) = \triangledown_{\alpha} \ell_{TSE} \left(w, \alpha \right)  + \triangledown_{w} \ell_{train} \left(w, \alpha \right)$
		\ENDFOR
	\ENDFOR
 	\STATE Derive the final architecture based on the learned $\alpha$
	\end{algorithmic}
\end{algorithm}

\newpage
\bibliographystyle{plain}
\bibliography {library}

%
%
\end{document}